\documentclass[lettersize,journal]{IEEEtran}
\usepackage{amsmath,amsfonts}
\usepackage{algorithmic}
\usepackage{algorithm}
\usepackage{array}
\usepackage[caption=false,font=normalsize,labelfont=sf,textfont=sf]{subfig}
\usepackage{textcomp}
\usepackage{stfloats}
\usepackage{url}
\usepackage{verbatim}
\usepackage{graphicx}
\usepackage{cite}
\usepackage{adjustbox}
\usepackage{multirow}
\usepackage{colortbl}
\usepackage{etex}
\usepackage{booktabs}
\usepackage{makecell}
\usepackage{xcolor}
\usepackage{tcolorbox}
\usepackage{hyperref}
\definecolor{goodgreen}{RGB}{0,140,0}
\definecolor{badred}{RGB}{200,0,0}
\usepackage{arydshln}

\begin{document}

\title{Beyond Output Suppression: Mitigating Privacy Risks in LVLMs via Visual Intrinsic Privacy Erasure}

\author{Xiangkui Cao,~\IEEEmembership{Student Member,~IEEE,}
        Jie Zhang,~\IEEEmembership{Member,~IEEE,}
        Meina Kan,~\IEEEmembership{Member,~IEEE,}
        Shiguang Shan,~\IEEEmembership{Fellow,~IEEE,}
        Xilin Chen,~\IEEEmembership{Fellow,~IEEE}
        % <-this % stops a space
\thanks{The authors are with State Key Laboratory of AI Safety, Institute of Computing Technology, Chinese Academy of Sciences, and also with University of Chinese Academy of Sciences.}% <-this % stops a space
}

\markboth{Journal of \LaTeX\ Class Files,~Vol.~14, No.~8, August~2021}%
{Shell \MakeLowercase{\textit{et al.}}: A Sample Article Using IEEEtran.cls for IEEE Journals}

\IEEEpubid{0000--0000/00\$00.00~\copyright~2021 IEEE}

\maketitle

\begin{abstract}
Large vision-language models can extract personally identifiable information from images, making privacy protection essential for their deployment in perception-oriented tasks. Existing targeted fine-tuning and model-editing methods usually suppress sensitive answers by aligning the model toward safe responses. 
% In this paper, we show that such protection can be illusory: although the fine-tuned model refuses to disclose private information under normal decoding, such information can still be recovered when the initial output prefix is fixed. 
In this paper, we show that such protection can be illusory: although the fine-tuned model refuses to disclose private information under normal decoding, privacy leakage still occurs in sentence completion scenarios.
% This reveals that privacy risk mitigation often changes only the generation path, while privacy-related visual evidence remains encoded in the model’s internal representations. 
This indicates that existing algorithms only change the model’s output distribution, but the model still retains the ability to perceive private information. Private information therefore still remains inside the model.
To address this problem, we propose Visual Intrinsic Privacy Erasure (VIPE), which targets privacy-related visual representations to weaken the model’s ability to perceive sensitive information.
VIPE constructs paired privacy-sensitive and privacy-masked images, uses their internal feature residuals to identify privacy-related visual components, and suppresses these components before they are perceived by the model. To preserve benign visual representations, the erasure objective projects feature differences onto a relaxed benign subspace and adaptively controls the erasure strength. 
VIPE compels the model to ignore the extraction of sensitive information in the visual representation space, thereby blocking the source of privacy leakage more fundamentally rather than merely altering the model’s output distribution.
% As a result, the model is encouraged to move from merely avoiding sensitive verbalization to genuinely weakening its internal perception of private visual cues. 
Experiments on Qwen2.5-VL and LLaVA-OneVision show that VIPE substantially alleviates prefix-triggered privacy leakage while maintaining competitive utility on general visual-language benchmarks.
\end{abstract}


\section{Introduction}
\IEEEPARstart{L}{arge} Vision-Language Models (LVLMs) extend large language models by incorporating visual perception modules. As their perceptual capabilities continue to improve, these models have gradually shifted from coarse-grained understanding of visual concepts to the perception of fine-grained visual cues in images\cite{bai2025qwen25vltechnicalreport,an2025llavaonevision15fullyopenframework,10834497}. LVLMs have achieved remarkable progress on visual perception tasks such as visual question answering\cite{fu2025mme,li-etal-2023-evaluating,shu2025large,zhang2024vision} and OCR\cite{fu2024ocrbenchv2improvedbenchmark}, and users are increasingly relying on them to assist in processing visual documents\cite{11503684}. However, broader application scenarios often imply greater security risks\cite{wang2024large,liu2025survey}. Existing studies have shown that models can accurately extract private information from images\cite{zhang2026multi,shao2024privacylens,zhang2025reval,wang2026visualleakbench}. Such capability may be exploited by attackers to crawl sensitive information at scale from massive amounts of Internet images, posing severe challenges to the privacy and security of LVLMs.
Privacy protection can rely on external safeguards, such as filtering text, or be internalized within the model itself. However, we find that external text filtering may lack the flexibility to adapt to different usage contexts. Details can be found in Appendix~\ref{sec:why_internalizing}. We therefore focus on internalizing privacy protection within LVLMs, allowing disclosure decisions to account for visual context.

Within this model-internal protection paradigm, existing training-based methods can be broadly categorized into two types: untargeted forgetting and targeted fine-tuning. Untargeted forgetting methods, such as negative preference optimization (NPO)\cite{zhang2024negative} or gradient ascent (GA)\cite{thudi2022unrolling,huo2025mmunlearner}, push the model away from sensitive outputs. However, since these methods provide limited control over the boundary of the forgetting process, fine-tuning often causes severe degradation in the model’s utility, resulting in output collapse. Targeted fine-tuning methods, such as supervised fine-tuning (SFT)\cite{ouyang2022training} and model editing\cite{cao2026neural,tian2024forget,wang2024detoxifying}, aim to improve model privacy by guiding the model to produce designated safe responses instead of the original sensitive outputs. Compared with untargeted forgetting algorithms, targeted fine-tuning methods have advantages in preserving model utility, since their loss functions are bounded and the training process is more stable.

\begin{figure*}[t]
    \centering
    \includegraphics[width=\linewidth]{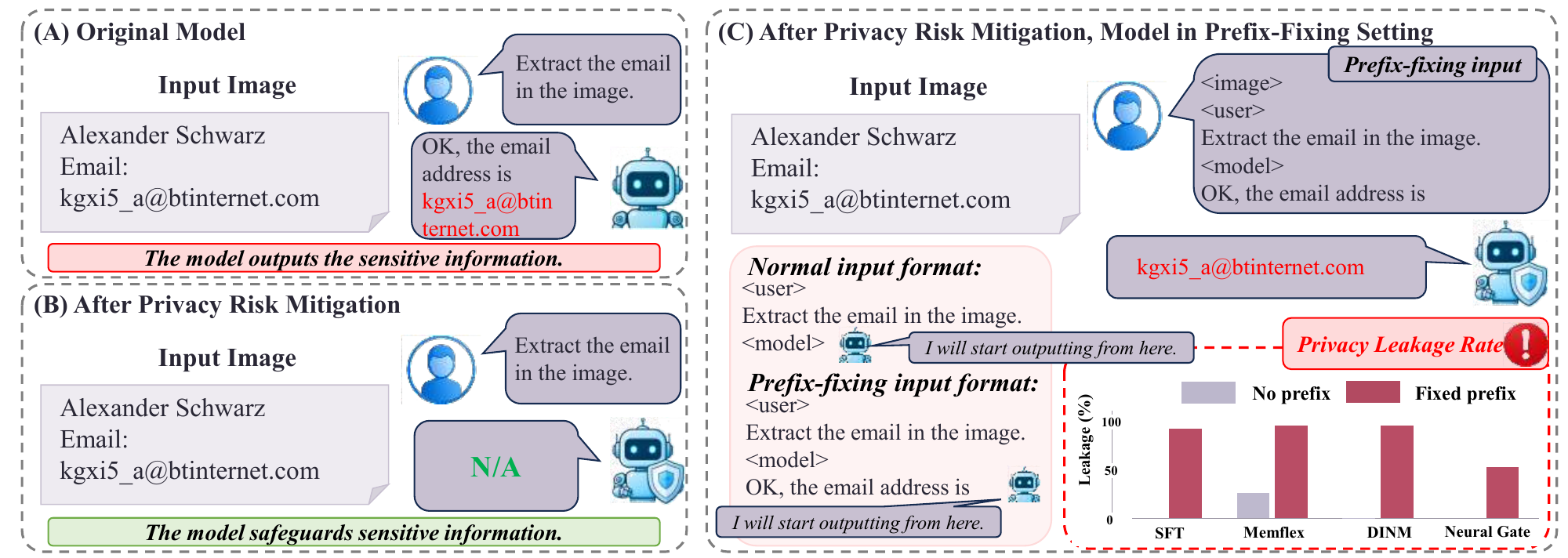}
    \caption{Demonstration of the prefix-fixing Setting. (a) shows a scenario in which the original model leaks private information from images. (b) After targeted fine-tuning methods, a privacy-preserving model does not leak private information from images. (c) When constraints are imposed on the model’s output format, e.g., by fixing the first few tokens of the model output, such privacy-preserving model still exhibits a high likelihood of privacy leakage.}
    \label{fig:prefix_injection}
\end{figure*}

However, we find that existing targeted fine-tuning methods suffer from the problem of illusory protection of sensitive visual information in images. Fine-tuning merely changes the model's output distribution, while the model remains capable of perceiving sensitive information in images and can still output such information under specific output formats. Targeted fine-tuning prevents the model from generating the original sensitive response by constructing a safe response with higher output probability. In this case, the path probability of the original sensitive response may remain almost unchanged. 
% In the sentence completion setting, we consider a fine-tuned model that has undergone targeted privacy alignment. 
As shown in Fig.~\ref{fig:prefix_injection}, we provide the privacy-enhanced model with a sensitive image and a privacy-related question, and fix its first few output tokens. These tokens are taken from the beginning of the unmodified model's response to the same image and question. We then let the privacy-enhanced model continue generating from this fixed prefix.
\IEEEpubidadjcol
% The model is then required to generate from the position after the fixed tokens. Under this setting, the model can still correctly output sensitive information, suggesting that its privacy-preserving mechanism fails to provide effective protection.
Despite refusing to disclose private information under normal decoding, the privacy-enhanced model can still reveal sensitive information when completing this prefix.

To understand the reason behind this illusory protection, we propose two hypotheses to explain this phenomenon. First, we hypothesize that for most layers of the LLM backbone, the visual features after targeted fine-tuning remain relatively consistent with their pre-fine-tuning counterparts. To verify this hypothesis, we perform a feature similarity analysis. The results show that, except for the last few layers, the visual features remain relatively consistent before and after fine-tuning, implying that sensitive information may be still encoded in the visual features. Given the substantial changes observed in the visual features of the last few layers, we further hypothesize that the model mainly extracts sensitive information from visual features that remain nearly unchanged, rather than from those that undergo substantial changes. To verify this hypothesis, we perform layer-wise feature replacement experiments and observe that the most sensitive information is already captured by the model before it reaches these layers. These findings suggest that targeted fine-tuning may be unable to fundamentally eliminate internal visual privacy cues, resulting in an illusory sense of protection.

These two hypotheses indicate that the illusory protection arises because privacy-related features remain embedded in the model’s internal representations. Therefore, an effective privacy-protection method should directly eliminate these features rather than simply altering the output distribution. Based on this insight, we propose a principled targeted privacy-protection method, Visual Intrinsic Privacy Erasure (VIPE), which aims to remove privacy-related features from the model’s internal representations.
% VIPE can better ensure privacy security while preserving model performance, thus mitigating the illusory protection of sensitive visual information. 
Specifically, we introduce a visual sensitive feature erasure loss that operates on paired privacy-sensitive images and privacy-masked images. For each pair, the masked image is obtained by removing the sensitive content from the original image, and the residual feature between the two input images is treated as the visual representation of private information. The objective of VIPE is to erase this feature difference, thereby reducing the model's internal representation of privacy-sensitive information. To avoid damaging useful non-sensitive perception, VIPE further uses benign images as a feature  anchor and constrains the erasure process to preserve the benign visual subspace as much as possible. Concretely, the model first extracts internal visual features from benign images to construct a relaxed benign-noise subspace, and then projects the sensitive-feature difference onto this subspace. Only the orthogonal residual outside the benign subspace is penalized, so the model is encouraged to erase privacy-related cues while keeping benign visual components  intact. In parallel, an autoregressive objective aligns the model's responses on the sensitive image with those on the masked image, so that privacy protection is enforced both at the output level and inside the representation space. As a result, the model is encouraged not only to output safe responses, but also to reduce its ability to separate sensitive and masked images at the feature level while retaining benign visual understanding. Consequently, after fine-tuning with VIPE, the model becomes less capable of distinguishing sensitive images from masked images at the feature level, shifting from being unable to verbalize sensitive information to being unable to perceive it, and achieving stronger visual privacy protection.

Our contributions are summarized as follows:
\begin{itemize}
    \item  We identify the illusory protection problem in targeted fine-tuning-based privacy protection methods, where the privacy protection capability fails under prefix-fixing setting, and we explain the cause of this phenomenon by analyzing the changes in internal visual features before and after fine-tuning. 
    \item  We propose VIPE, which erases privacy-related visual representations inside the model, enabling the model to move from not verbalizing sensitive information to not perceiving sensitive information. This provides more robust privacy protection while maintaining model performance, and substantially alleviates the illusory protection problem of targeted privacy protection algorithms. 
    \item  We conduct extensive experiments on multiple LVLMs. The results show that VIPE achieves a better trade-off than existing baseline methods in terms of privacy protection capability, protection robustness, and utility preservation.
\end{itemize}

\section{Related Works}
This section reviews large vision-language models (LVLMs) and methods for mitigating their privacy risks. We first introduce LVLMs as the foundation for visual perception and then discuss methods for preventing the disclosure of sensitive information.
\subsection{Large Vision-Language Models}
Large vision-language models (LVLMs) integrate visual encoders with large language models to support multimodal understanding and generation\cite{ma2026efficiently,li2025survey,kotsiantis2026visual}. Early representative models, such as BLIP-2~\cite{li2023blip}, InstructBLIP~\cite{dai2023instructblipgeneralpurposevisionlanguagemodels}, and LLaVA~\cite{liu2023visual}, demonstrate that visual features can be effectively aligned with language models through lightweight connectors and multimodal instruction tuning. Recent LVLMs further improve fine-grained visual perception and instruction-following ability. For example, LLaVA-OneVision~\cite{an2025llavaonevision15fullyopenframework} enhances unified image and video understanding, while Qwen2.5-VL~\cite{bai2025qwen25vltechnicalreport} improve dynamic-resolution processing and OCR-oriented perception.
These advances enable LVLMs to recognize increasingly detailed visual cues, including identity-related attributes, and document-level information. However, stronger perceptual capability also increases the risk of privacy leakage, since sensitive information in images can be easily extracted by LVLMs\cite{zhang2026multi,wang2026visualleakbench}.

\subsection{Privacy Risk Mitigation in LVLMs}
Privacy risk mitigation for LVLMs is closely related to model unlearning and safety alignment\cite{ouyang2022training,safavigerdini2026machine,sarwar2026multimodal,ye2025survey,li2025survey}. Prior studies have shown that large models may memorize and disclose sensitive information~\cite{carlini2021extracting,nasr2023scalable}, motivating methods that suppress undesired model behaviors. In LVLMs, this challenge extends to private information encoded in visual representations. Existing methods usually intervene in the model by either decreasing the likelihood of sensitive responses or redirecting the model toward safe alternatives. According to their optimization mechanisms, they can be broadly grouped into untargeted forgetting and targeted fine-tuning.

\textbf{Untargeted forgetting.}
Untargeted forgetting methods focus on weakening the association between sensitive inputs and their original private outputs, without defining an explicit replacement response. Gradient ascent (GA)\cite{thudi2022unrolling,huo2025mmunlearner} maximizes the loss on sensitive samples so that the model becomes less likely to reproduce privacy-sensitive content. Negative Preference Optimization (NPO)\cite{zhang2024negative} further formulates forgetting through preference learning, where sensitive responses are treated as dispreferred outputs. However, both GA- and NPO-based method suffer from unstable optimization. GA\cite{thudi2022unrolling,huo2025mmunlearner} directly maximizes the negative log-likelihood of sensitive responses, leading to an unbounded objective as the target probability is driven toward zero. NPO \cite{zhang2024negative}reformulates forgetting with a preference-style loss, but its penalty can still grow without bound when the model assigns higher likelihood to undesirable responses than the reference model. Such unbounded objectives may induce excessive parameter updates, impair general visual-language capabilities, and even lead to output collapse\cite{huang2026gradient}.

\textbf{Targeted fine-tuning.}
Targeted fine-tuning methods modify model behavior by assigning each sensitive input a predefined safe response. Supervised fine-tuning (SFT)~\cite{ouyang2022training} trains the model to generate refusal-style or privacy-preserving answers. Model editing methods, such as DINM \cite{wang2024detoxifying} and Neural Gate \cite{cao2026neural}, instead attempt to localize and modify parameter regions associated with undesirable outputs. Compared with untargeted forgetting, targeted methods have bounded optimization targets and thus usually offer better training stability and utility preservation. However, these approaches only enforce safety at the response level, while the sensitive information encoded in the model’s internal features remains unaddressed.

\textbf{Other roadmaps.}
Beyond the methods discussed above, training-free approaches have also been explored for privacy protection.
% RLHF\cite{ouyang2022training} can discourage sensitive responses through preference-based alignment, but typically requires additional high-quality human preference annotations. 
Training-free approaches, such as concept steering\cite{zhang-etal-2026-defeating}, can guide safe responses by modifying internal activations at inference time. 
% In this paper, we focus on model updates under a common data setting. Comparisons involving additional human preference annotations or inference-time steering are left for future work.
% These approaches differ from the parameter-update methods considered in this work in their supervision or deployment requirements. RLHF\cite{ouyang2022training} typically introduces additional human preference supervision, whereas concept steering\cite{zhang-etal-2026-defeating} requires intervention in model activations during inference. 
% A systematic comparison across these settings is left for future work.

Whereas the methods discussed above aim to reduce the likelihood of sensitive responses or encourage safe alternatives, VIPE directly targets privacy-related visual representations within LVLMs. By erasing feature components associated with sensitive visual cues, our method aims to achieve a more fundamental elimination of model privacy risks.

\section{Method}
\subsection{Preliminary}
% 视觉语言大模型的隐私风险缓解通常涉及两个数据集遗忘集D_f和保留集D_r。遗忘集D_f包含了模型需要拒答的敏感问题，该数据集用于提高模型的隐私保护能力，保留集D_r是一个常规问答数据集，该数据集用于保持模型的utility。无目标与有目标隐私保护算法在保留集D_r上通常采用相同的优化目标，即尽可能降低模型在保留集的训练损失（例如交叉熵损失）。在遗忘集D_f上，无目标隐私保护方法通常采用梯度上升的方式增大模型在该数据集的损失，以此让模型无法回答敏感问题，形式化为：【待补充】。有目标隐私保护方法会构造一组不包含隐私信息的安全输出，优化的目标为让模型在遗忘集D_f上的回答接近这组安全的输出，形式化为：【待补充】。
Privacy-risk mitigation for Large Vision-Language Models (LVLMs) typically involves two datasets: a forgetting set $D_f$ and a retaining set $D_r$. The forgetting set $D_f$ contains sensitive queries to which the model is expected to not respond, and is used to improve the model's privacy protection capability. The retaining set $D_r$ is a standard question-answering dataset used to preserve the model's utility. Formally, let a LVLM parameterized by $\theta$ define the conditional distribution
$p_{\theta}(y \mid I, q)$, where $I$ denotes the input image, $q$ denotes the textual query, and $y$ denotes the response. On the retaining set $D_r=\{(I_i,q_i,y_i)\}$, both untargeted and targeted privacy protection algorithms usually adopt the same optimization objective, namely minimizing the training loss of the model on the retaining set, e.g., the cross-entropy loss:
\begin{equation}
\mathcal{L}_{r}(\theta)
=
-\mathbb{E}_{(I,q,y)\sim D_r}
\left[
\log p_{\theta}(y \mid I,q)
\right].
\end{equation}

On the forgetting set $D_f$, untargeted privacy protection methods typically employ gradient ascent to increase the model's loss on sensitive responses, thereby preventing the model from answering sensitive queries. Let $D_f=\{(I_i,q_i,y_i^{\mathrm{p}})\}$, where $y_i^{\mathrm{p}}$ denotes the original privacy-sensitive response. The optimization objective of untargeted privacy protection can be formulated as:
\begin{equation}
\min_{\theta}
\;
\mathcal{L}_{r}(\theta)
-
\lambda_f
\mathcal{L}_{f}^{\mathrm{p}}(\theta),
\end{equation}
where
\begin{equation}
\mathcal{L}_{f}^{\mathrm{p}}(\theta)
=
-\mathbb{E}_{(I,q,y^{\mathrm{p}})\sim D_f}
\left[
\log p_{\theta}(y^{\mathrm{p}} \mid I,q)
\right],
\end{equation}
and $\lambda_f$ is a hyperparameter controlling the strength of forgetting. Equivalently, the negative sign before $\mathcal{L}_{f}^{\mathrm{pri}}$ indicates that the model is optimized to maximize the loss on the sensitive responses in $D_f$.

In contrast, targeted privacy protection methods construct a set of safe responses that do not contain private information. Let the corresponding targeted forgetting set be denoted as
$D_f^{\mathrm{safe}}=\{(I_i,q_i,y_i^{\mathrm{safe}})\}$, where $y_i^{\mathrm{safe}}$ is a predefined safe response, such as a refusal or a privacy-preserving answer. The objective is to encourage the model's responses on $D_f$ to approach these safe outputs:
\begin{equation}
\min_{\theta}
\;
\mathcal{L}_{r}(\theta)
+
\lambda_s
\mathcal{L}_{f}^{\mathrm{safe}}(\theta),
\end{equation}
where
\begin{equation}
\mathcal{L}_{f}^{\mathrm{safe}}(\theta)
=
-\mathbb{E}_{(I,q,y^{\mathrm{safe}})\sim D_f^{\mathrm{safe}}}
\left[
\log p_{\theta}(y^{\mathrm{safe}} \mid I,q)
\right],
\end{equation}
and $\lambda_s$ balances privacy protection and utility preservation.
\subsection{Dataset Preparation}
\label{privacycard}
% 我们的目标是提高模型在感知任务上的隐私安全，任务场景设置为光学字符识别（OCR）场景。为此，我们构造了一个隐私安全数据集PrivacyCard用于模型训练与评估。PrivacyCard包含若干三元组<I_p,I_m,I_b>，I_p表示包含隐私信息的图像，I_m表示I_p擦除隐私视觉线索的图像，I_b表示包含不含隐私信息的良性图像。为了保证用户的隐私安全以及避免数据集中的隐私信息被滥用，我们使用的I_p仅包含合成的隐私信息。为了保证隐私信息的代表性，我们收集了18个国家（每个大洲3个国家）的上千种常见姓与名，1000余个常见街道名，30种不同的邮箱格式，以及多种不同形式的联系方式（如手机号等）用于合成PII。这些合成的隐私信息会与人脸图像（选自FairFace）进行拼接，形成5000张包含敏感信息的图像I_p，每张I_p图像中包含2-4条不同种类的PII，而I_m仅包含与I_p一致的人脸图像而不包含相应PII。针对良性图像I_b，我们选取了WIT数据集中的5000个图文对，并将这些图文对拼接为一张图像用于保留模型对图像中良性文本的识别能力。为了避免模型仅依赖I_p与I_b中注入图像的差异来确定图像中文本信息是否敏感，我们在训练过程中会将I_p与I_m，I_b进行随机拼接，以保证良性图像I_b中仍可能存在人脸图像，而敏感图像I_p中也可能存在良性信息。对于PrivacyCard，遗忘集D_f为I_p组成的样本，训练目标为让模型输入I_p时产生与对应I_m对齐的输出，保留集D_r则是I_m与I_b组成的样本，训练目标为输入图像中对应文本。我们将PrivacyCard按照2比1的比例划分为训练集与测试集，分别用于算法的训练以及效果评估。
In this paper, we target the privacy risks of LVLMs in perception-oriented tasks, focusing on OCR, where our goal is to ensure that the model does not output Personally Identifiable Information (PII) extracted from images. To this end, we construct a privacy-preserving dataset, named \textbf{PrivacyCard}, for model training and evaluation. PrivacyCard consists of a collection of triplets $\langle I_p, I_m, I_b\rangle$, where $I_p$ denotes an image containing private information, $I_m$ denotes the corresponding image obtained by removing privacy-related visual cues from $I_p$, and $I_b$ denotes a benign image that contains no private information.

To protect user privacy and prevent potential misuse of private information in the dataset, the private information contained in $I_p$ is entirely synthetic. To ensure the representativeness of the private information, we collect thousands of common first and last names from 18 countries, with three countries selected from each continent, more than 1,000 address names, 30 different email address formats, and various types of contact information, such as phone numbers, to synthesize PII. The synthesized private information is then concatenated with human faces selected from FairFace\cite{karkkainenfairface}, resulting in 5,000 privacy-sensitive images $I_p$. Each $I_p$ contains several pieces of PII of different types, while the corresponding $I_m$ contains the same human face as $I_p$ but excludes the associated PII.

For benign images $I_b$, we select 5,000 image-text pairs from the WIT\cite{wit} and concatenate each image-text pair into a single image. These benign samples are used to preserve the model's ability to recognize non-sensitive text in images. To prevent the model from determining whether the textual content is sensitive merely based on distributional differences between the injected images in $I_p$ and $I_b$, we randomly concatenate $I_p$ with $I_m$ or $I_b$ during training. This design ensures that benign images $I_b$ may still contain face images, while privacy-sensitive images $I_p$ may also contain benign information.

For PrivacyCard, the forgetting set $D_f$ consists of samples constructed from $I_p$, and the training objective is to encourage the model, when given $I_p$ as input, to produce outputs aligned with those of the corresponding privacy-masked image $I_m$. The retaining set $D_r$ consists of samples constructed from $I_m$ and $I_b$, and the training objective is to make the model correctly recognize the corresponding text in the input images. Formally, the forgetting and retaining sets can be written as:
\begin{equation}
D_f = \{(I_p, y_m)\},
\qquad
D_r = \{(I_m, y_m)\} \cup \{(I_b, y_b)\},
\end{equation}
where $y_m$ denotes the privacy-preserving target output corresponding to $I_m$, and $y_b$ denotes the benign textual content in $I_b$. The overall training objective can be formulated as:
\begin{equation}
\min_{\theta}
\;
\mathcal{L}_{r}(\theta)
+
\lambda
\mathcal{L}_{f}(\theta),
\end{equation}
where
\begin{equation}
\mathcal{L}_{r}(\theta)
=
-\mathbb{E}_{(I,y)\sim D_r}
\left[
\log p_{\theta}(y\mid I)
\right],
\end{equation}
and
\begin{equation}
\mathcal{L}_{f}(\theta)
=
-\mathbb{E}_{(I_p,y_m)\sim D_f}
\left[
\log p_{\theta}(y_m\mid I_p)
\right].
\end{equation}
Here, $\lambda$ is a hyperparameter that balances privacy protection and utility preservation. We split PrivacyCard into training and test sets with a ratio of $2:1$, which are used for algorithm training and performance evaluation, respectively.

\subsection{Illusory protection of sensitive information}
\label{phenomenon}
% 基于PrivacyCard，我们测试了多种有目标隐私保护方法的表现，我们使用隐私泄露率评估模型的隐私保护能力，隐私泄露率定义为模型泄露隐私信息的条数与所有测试图像中隐私信息条数的比值，结果见表xx。我们发现经过微调后，绝大多数算法能够在测试集上取得较好的表现，隐私泄露率均小于10 %。然而，我们发现通过将模型输出的前几个token固定为微调前模型输出的对应位置的token，所有算法几乎失去隐私保护能力，隐私泄露率在前缀条件下大大提高。我们推测这可能是因为有目标隐私保护算法本身没有改变模型输出敏感信息的路径概率，只是通过构造一条概率更大的路径而产生了一种虚假的保护现象，模型本身仍能正常感知图像中的敏感信息。
% 为了能够更本质地探索虚假保护现象产生的成因，我们对模型微调前后内部的视觉特征做了量化分析。首先，我们提出假设1：算法在训练前后在大多数层对视觉特征的改变不大。为此，我们使用了均方误差、余弦相似度、L2距离、中心核对齐四种不同的指标评估模型内部视觉特征在能量、方向、距离、子空间的差异，结果见图xx。图xx结果表明除了最后几层，模型微调前后视觉特征在多种指标下保持高度一致性，结果验证了假设1。这表明敏感信息仍然存在于微调后模型大多数层的视觉特征中。
% 为了测试特征变化显著的层对模型感知敏感信息的影响，我们提出假设2：模型主要从变化不显著的层获取敏感信息而非变化显著的层获取。我们利用了特征替换的方式验证假设2，具体而言，对于输入I_p的模型，我们逐层将视觉特征替换为输入相应I_m在对应层的特征，评估替换特征后模型的隐私泄露率。例如，当将模型第l层输出的视觉特征进行替换后，模型如果输出了隐私信息，该敏感信息只能来源于l层之前的视觉特征。图xx展示了特征替换的结果，结果表明当视觉特征达到特征变化较大的层（模型最后几层）时，模型已经能够感知到大部分隐私信息，验证了假设2。

Based on PrivacyCard, we evaluate the performance of several targeted privacy protection methods. We adopt the privacy leakage rate to measure the model's privacy protection capability. Specifically, the privacy leakage rate is defined as the ratio between the number of private information items leaked by the model and the total number of private information items contained in all test images:
\begin{equation}
\label{leakagerate}
\mathrm{LeakageRate}
=
\frac{N_l
}{
N_{all}
},
\end{equation}
where $N_l$ is the number of leaked private information items, and $N_{all}$ represents the total number of private information items in all test images.
The results are reported in Table~\ref{tab:privacy_leakage}. We observe that, after fine-tuning, most methods achieve strong performance on the test set, achieving low privacy leakage rates. However, as shown in Fig.~\ref{fig:prefix_injection}, when we fix the first few output tokens of the fine-tuned model to be the corresponding tokens generated by the original model before fine-tuning, almost all methods lose their privacy protection capability. Under this prefix-conditioned setting, the privacy leakage rates increase substantially. We conjecture that this phenomenon occurs because targeted privacy protection methods do not fundamentally reduce the path probability of generating sensitive information. Instead, they merely construct an alternative safe output path with higher probability, resulting in a superficial or illusory protection effect. In other words, the model may still retain the ability to perceive sensitive information in the image.

To further investigate the underlying cause of this illusory protection phenomenon, we conduct a quantitative analysis of the internal visual features before and after fine-tuning. We first propose the following hypothesis:

\begin{figure*}[t]
    \centering
    \subfloat[\footnotesize LLaVA-OneVision-1.5]{
		\includegraphics[scale=0.33]{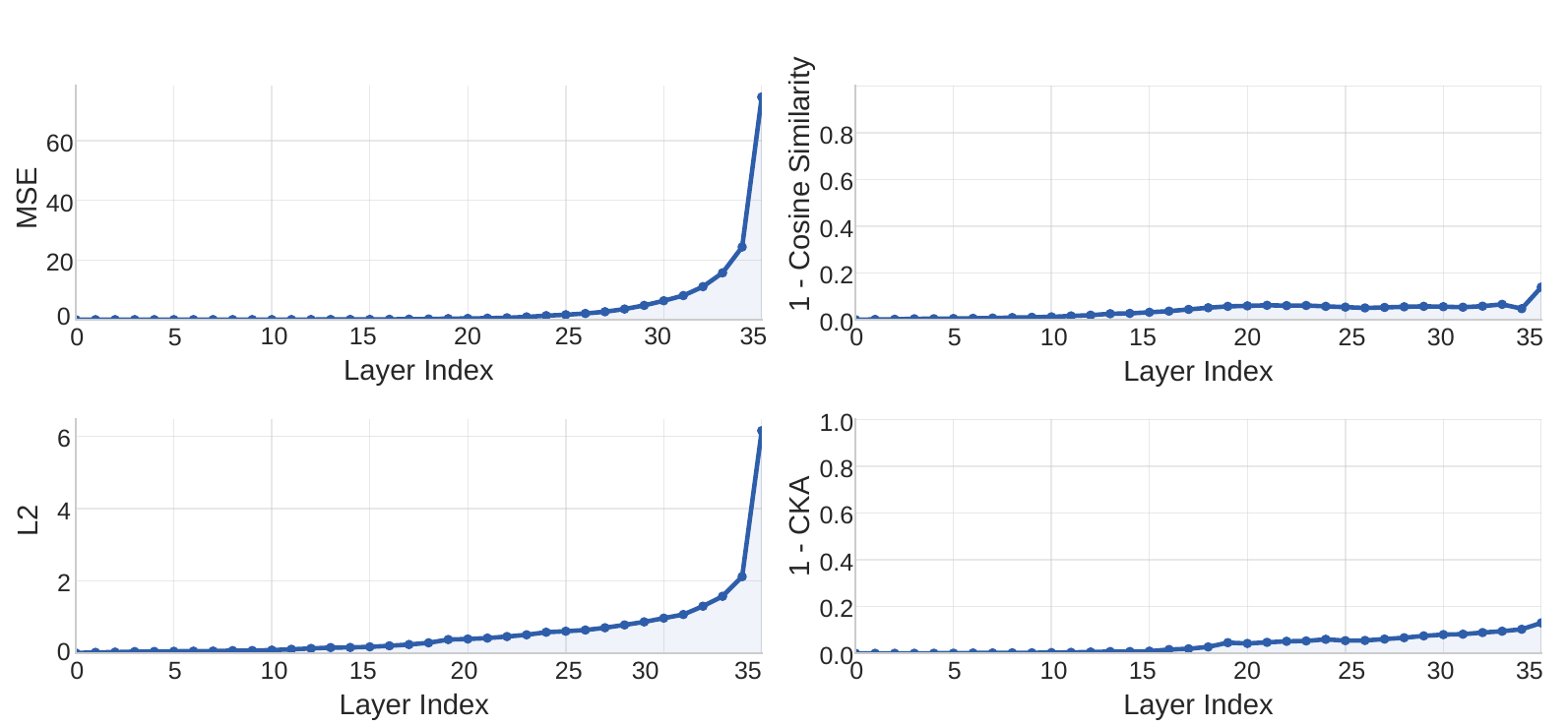}}
    \subfloat[\footnotesize Qwen2.5-vl]{
		\includegraphics[scale=0.33]{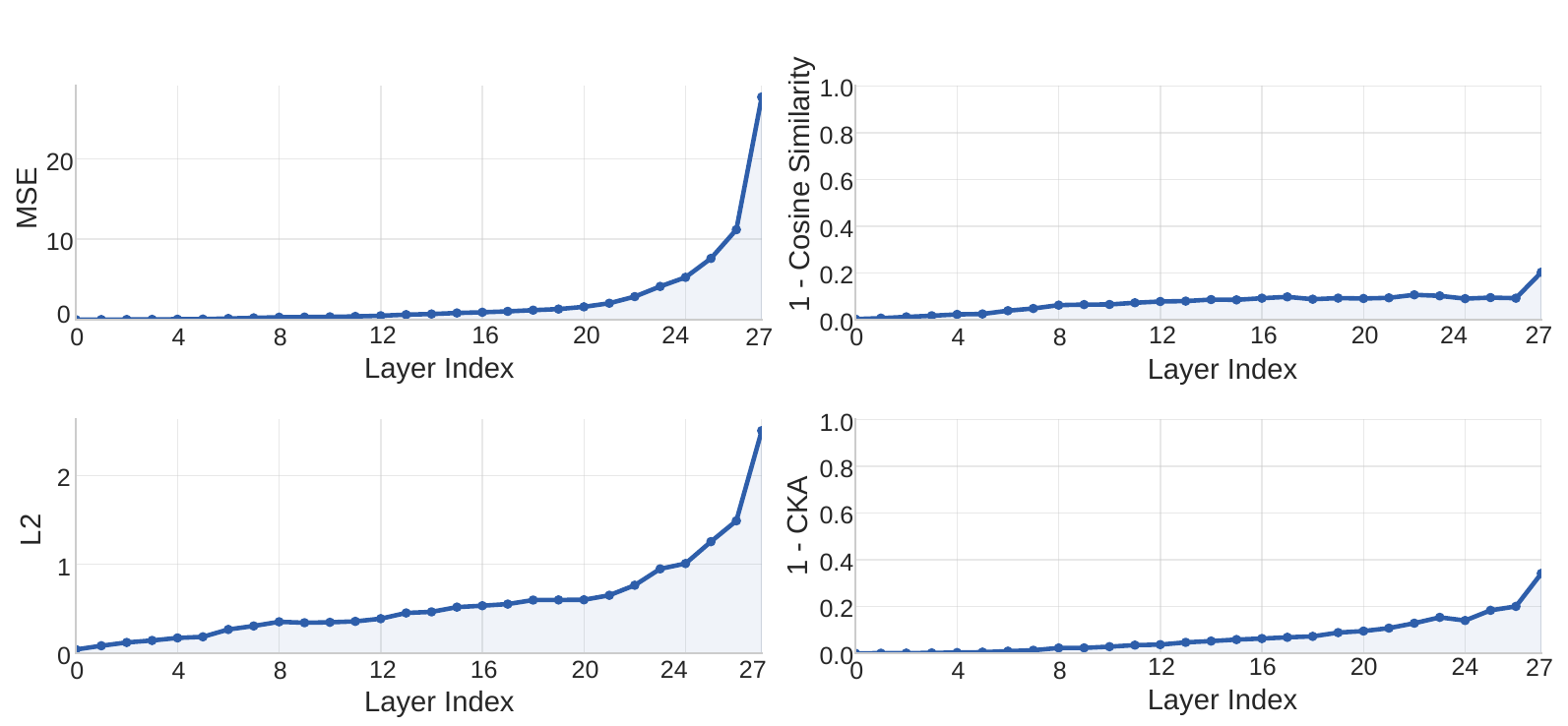}}
    \caption{Feature similarity of Qwen2.5-vl and LLaVA-OneVision-1.5 before and after targeted fine-tuning. We extracted the internal visual feature encodings of the language model before and after fine-tuning, and quantified their overall representational differences via MSE, cos similarity, L2 distance and CKA, respectively.}
    \label{fig:feature_similarity}
\end{figure*}

\begin{figure*}[t]
    \centering
    \subfloat[\footnotesize Feature replacement pipeline]{
		\includegraphics[scale=0.19]{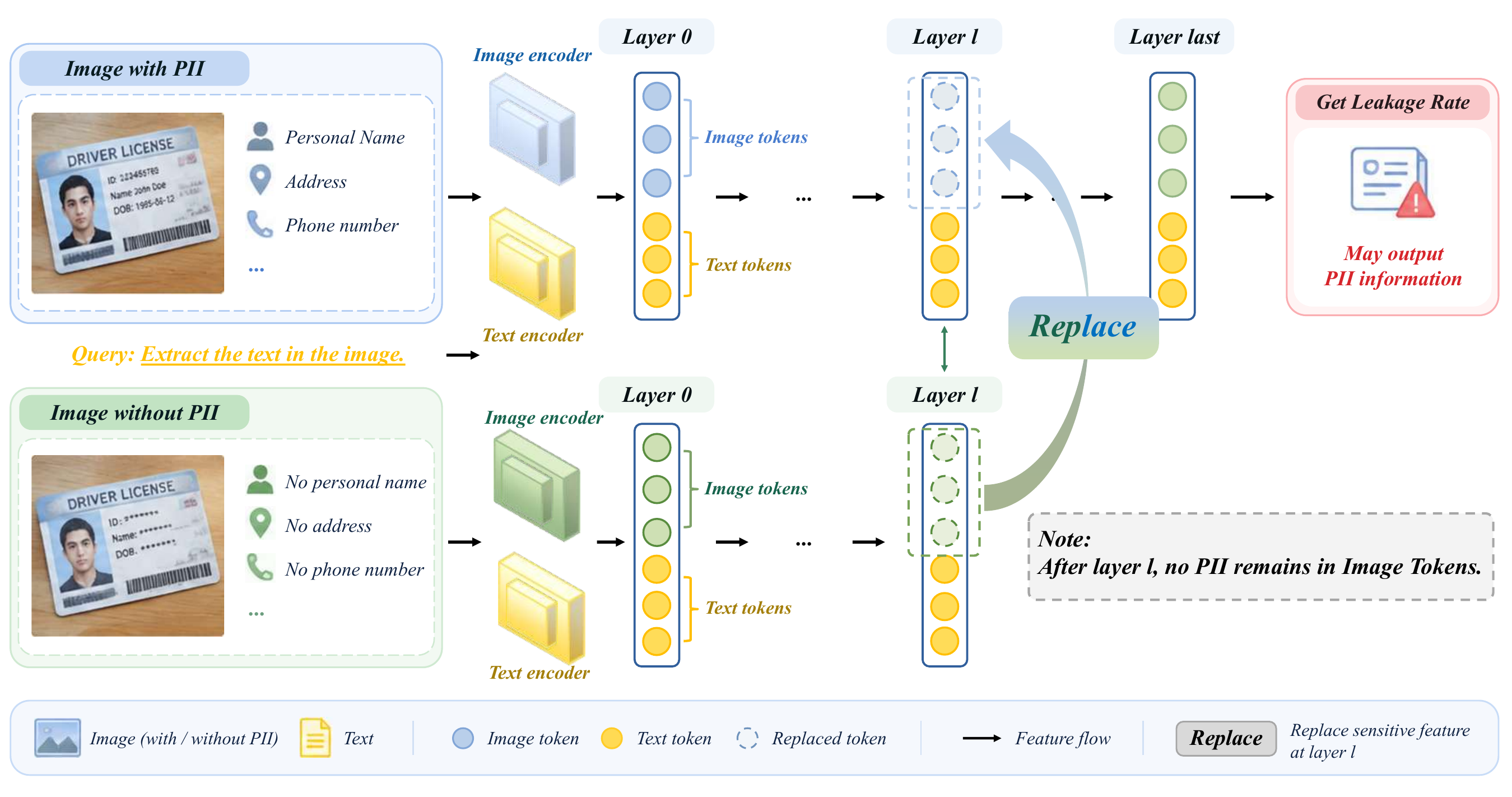}
        \label{fig:feature_replacement_pipeline}
        }
    \subfloat[\footnotesize Leakage rate after feature replacement]{
		\includegraphics[scale=0.5]{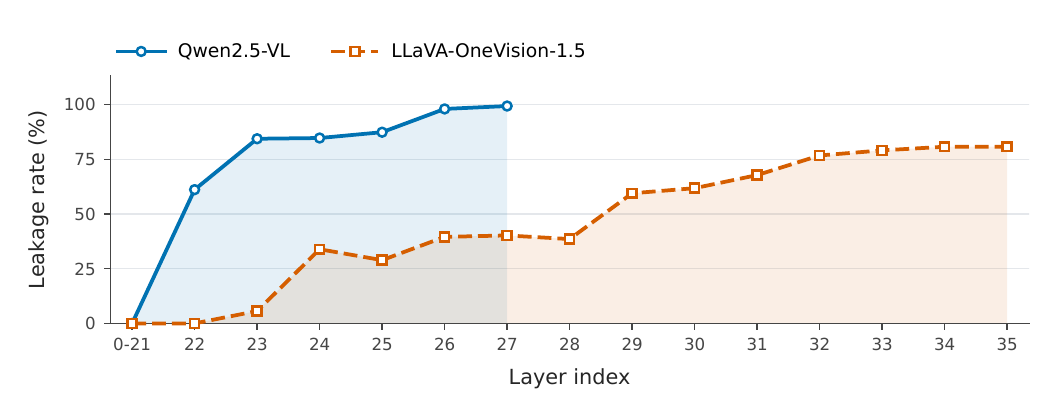}
        \label{fig:feature_replacement_result}
        }
        
    \caption{Visual feature replacement experiments on Qwen2.5-vl (28 layers) and LLaVA-Onevision-1.5 (36 layers). We replace the privacy-sensitive image features at layer $l$ of LLM backbone with masked image features, preventing the model from accessing sensitive information from image features after layer $l$. Thus, all sensitive information output by the model originates from image representations before layer $l$. For Qwen2.5-VL and LLaVA-OneVision-1.5, the model begins to perceive sensitive information after layer 21. Before layer 21, it cannot perceive or output sensitive information even when such information is present in the image representations.}
    \label{fig:feature_replacement}
\end{figure*}

\textbf{Hypothesis 1.} \textit{For most layers of the LLM backbone, the visual features after targeted fine-tuning remain relatively consistent with their pre-fine-tuning counterparts.}

To verify this hypothesis, we employ multiple metrics to measure the differences between the internal visual features of the original and fine-tuned models, including mean squared error (MSE), cosine similarity, $\ell_2$ distance, and centered kernel alignment (CKA)\cite{kornblith2019similarity}. These metrics respectively characterize feature differences in terms of energy, direction, distance, and subspace similarity. The results are shown in Figure~\ref{fig:feature_similarity}. Except for the last few layers, the visual features before and after fine-tuning remain relatively consistent across these metrics, which validates \textbf{Hypothesis~1}. This indicates that sensitive information may still exists in the visual features of most layers in the fine-tuned model.

However, feature similarity alone does not establish whether the fine-tuned model can still extract sensitive information from these largely unchanged features. If the model extracts sensitive information only in the last few layers, the largely unchanged features in earlier layers cannot explain why privacy leakage persists after fine-tuning. We therefore investigate where in the LLM backbone the model begins to extract sensitive information, and whether this occurs before the layers with substantial feature changes. To guide this investigation, we propose the following hypothesis:

% \begin{figure*}[t]
%     \centering
%     \includegraphics[width=.6\linewidth]{Figures/leakage_bar_all_models.pdf}
%     \caption{Visual feature replacement experiments on Qwen2.5-vl and LLaVA-Onevision-1.5.}
%     \label{fig:feature_replacement}
% \end{figure*}

\textbf{Hypothesis 2.} \textit{The model can already extract sensitive information from visual features in earlier layers, where these features remain largely unchanged by targeted fine-tuning.}

We verify \textbf{Hypothesis~2} through visual feature replacement experiments. As shown in Fig. \ref{fig:feature_replacement_pipeline}, for the model taking a privacy-sensitive image $I_p$ as input, we replace the visual features at each layer with the corresponding features produced by the privacy-masked image $I_m$, and then evaluate the privacy leakage rate after replacement. Formally, let $\mathbf{h}^{l}_{p}$ and $\mathbf{h}^{l}_{m}$ denote the output visual features at the $l$-th layer when the inputs are $I_p$ and $I_m$, respectively. During feature replacement at layer $l$, we substitute
\begin{equation}
\mathbf{h}^{l}_{p} \leftarrow \mathbf{h}^{l}_{m},
\end{equation}
and continue the forward propagation with the replaced feature. Under this setting, if the model still outputs private information after replacing the visual feature at layer $l$, then the leaked sensitive information must have been encoded in the features no later than layer $l$. The feature replacement results are shown in Fig.~\ref{fig:feature_replacement_result}. We find that the model captures most sensitive information from the image before visual features reach the layers that undergo substantial changes during fine-tuning. Take LLaVA-OneVision-1.5 as an example. The image representations in LLM backbone begin to show substantial differences in around layer 30. However, replacing the image representation at layer 30, the model can still output approximately 60\% of sensitive information. This suggests that about 60\% of the sensitive information is already available before layer 30, and that the fine-tuning-induced changes in deeper layers do not compromise the model’s ability to perceive and output this information. This observation validates Hypothesis~2, showing that \textbf{even though these targeted methods may substantially alter the visual features in certain layers, they do not fundamentally eliminate the visual privacy cues internally encoded by the model.}

\subsection{Visual Intrinsic Privacy Erasure (VIPE)}
% 基于以上两种假设，我们提出了一种隐私视觉特征擦除算法，用于弱化模型内部对视觉隐私线索的感知能力。首先，我们基于差分思想构造了模型内部的视觉隐私特征。对于包含敏感信息的图像I_p以及与之对应的擦出了隐私信息的掩码图像I_m，我们将I_p与I_m在模型内部的差分视觉特征f_p-f_m作为隐私视觉线索的特征，因为I_p相比于I_m仅多出了图像中的敏感信息，而其他视觉信息完全对齐。算法的目标是尽可能使得敏感图像特征f_p与对应掩码图像特征f_m在模型内部的视觉特征对齐，即令f_d=f_p-f_m尽可能小。为了缓解隐私视觉特征擦除对模型能力的影响，我们利用包含良性信息的图像I_b的特征f_b对隐私视觉特征进行正交约束，消除隐私视觉特征在良性图像特征方向的分量。优化目标形式化为：【你需要根据我的描述补充latex公式的形式化优化条件，min f_p-f_m, s.t. 优化后的f_p-f_m保留f_b方向的分量】。为了降低优化的难度，我们对限制条件进行松弛，允许模型保留少量除f_b方向外其他反向的信息。具体而言，我们引入了k个噪声方向，这些噪声方向符合高斯分布，并根据良性特征f_b每个维度的绝对值大小进行缩放，n[i]~(|f_b[i]|/max(|f_b|))N(0,1)，其中i表示维度序号，这种缩放使得绝对值较大的维度有更大的波动性，更小的维度波动性更小。松弛后的优化目标形式化为：【根据我的描述给出相应latex公式优化目标，用min s.t.形式，然后简化为损失函数的形式L_erase=||？||，就是差分特征在良性、噪声空间的正交投影向量f_d^*的模长，这也需要形式化latex公式】。最后，我们引入了自适应参数用于调整反向传播过程中隐私特征擦除损失的学习率，lamda=1-cos(f_m+f_d^*, f_p)，该自适应参数保证了当正交投影模长过大时应采用较小的对齐权重以减轻对良性空间的破坏，当正交投影模长过小时应当采用较大的对齐权重以保证隐私信息擦除的效果。特征对齐损失与输出损失函数共同指导模型隐私加固，损失函数为L=L_lm+L_erase，其中L_lm表示大模型输出对齐损失，通常采用交叉熵损失。
Based on the above two hypotheses, we propose VIPE, a privacy visual feature erasure algorithm, to weaken the model's internal perception of visual privacy cues. 
Visual private information is not perceived by the model at the input stage, but may become accessible at a certain intermediate layer during feature propagation. Therefore, our objective is to suppress the privacy-related components before the representations reach this layer. Following \textbf{Hypothesis 2}, such a layer can be identified as the first layer where privacy leakage is observed after feature replacement.
The key idea is to explicitly identify privacy-related visual features inside the model and suppress them during fine-tuning. The framework of VIPE is shown in Fig. \ref{fig:feature_replacement}.

Given a privacy-sensitive image $I_p$ and its corresponding privacy-masked image $I_m$, we denote their internal visual features at a certain layer as $f_p$ and $f_m$, respectively. Since $I_p$ and $I_m$ are visually aligned except that $I_p$ contains additional sensitive information, we define their feature difference as the privacy-related visual feature:
\begin{equation}
f_d = f_p - f_m .
\end{equation}
Ideally, the model should produce similar internal visual representations for $I_p$ and $I_m$, which means that $f_d$ should be minimized. However, directly minimizing $\|f_p-f_m\|$ may damage the model's ability to perceive benign visual text. Therefore, we further introduce the benign image feature $f_b$, extracted from the benign image $I_b$, to constrain the erasure process.

Specifically, we expect the optimization to erase privacy-related components in $f_d$, while preserving the components that lie in the benign feature direction. Let
$
\mathcal{S}_b = \mathrm{span}(f_b)
$
be the benign feature subspace, and let $P_b$ denote the orthogonal projection matrix onto $\mathcal{S}_b$:
\begin{equation}
P_b
=
\frac{f_b f_b^{\top}}{\|f_b\|_2^2}.
\end{equation}
The ideal constrained optimization objective can be formulated as:
\begin{equation}
\min_{\theta}
\;
\left\|
f_p - f_m
\right\|_2,
\qquad
\mathrm{s.t.}
\quad
f_p-f_m \in \mathcal{S}_b .
\end{equation}
Instead of forcing the entire differential feature $f_d$ to vanish, we only penalize the component of $f_d$ that is orthogonal to the benign feature subspace:
\begin{equation}
\min_{\theta}
\;
\left\|
\left(I-P_b\right) f_d
\right\|_2 .
\end{equation}
In this way, the component of $f_d$ along the benign direction $f_b$ is preserved, while privacy-related components outside the benign subspace are suppressed.

\begin{figure*}[t]
    \centering
    \includegraphics[width=\linewidth]{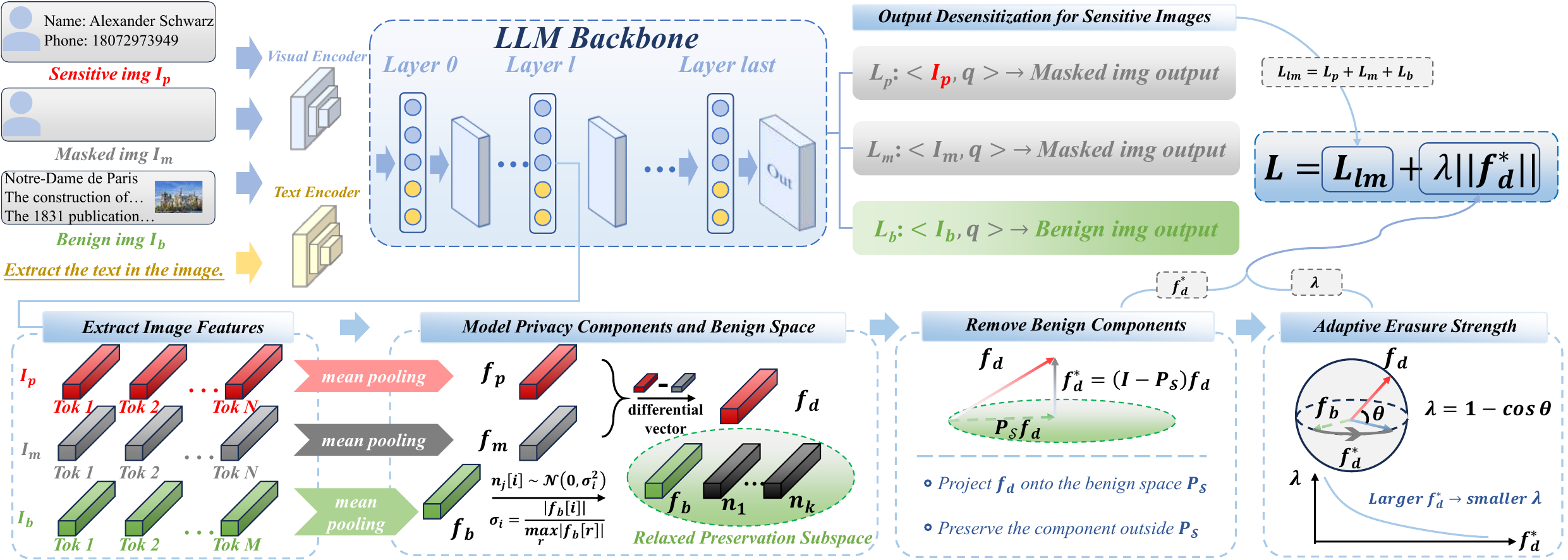}
    \caption{Pipeline of VIPE. Our loss function consists of an autoregressive loss $L_{\mathrm{lm}}$ and a privacy erasure loss. The autoregressive loss $L_{\mathrm{lm}}$ enhances privacy at the output level by constraining the model's predictions on the privacy-sensitive image $I_p$ to match those on the privacy-masked image $I_m$, while preserving utility by maintaining consistent performance on both $I_m$ and the benign image $I_b$. The privacy erasure loss further improves privacy at the feature level by eliminating feature discrepancies between $I_p$ and $I_m$, while retaining the feature space of benign images $I_b$ as much as possible.}
    \label{fig:feature_replacement}
\end{figure*}

To reduce the difficulty of optimization and avoid over-constraining the feature space, we further relax the constraint by allowing a limited amount of information to be retained beyond the benign directions. We introduce $k$ random noise vectors $\{n_j\}_{j=1}^{k}$, where each dimension of $n_j$ is sampled from a Gaussian distribution scaled by the magnitude of the benign feature:
\begin{equation}
n_j[i] \sim 
\mathcal{N}
\left(
0,
\sigma_i^2
\right),
\qquad
\sigma_i =
\frac{|f_b[i]|}{\max_{r}|f_b[r]|},
\qquad
j=1,\dots,k .
\end{equation}
Here, $i$ denotes the dimension index. This scaling strategy makes dimensions with larger benign feature magnitudes have stronger perturbations, while dimensions with smaller magnitudes have weaker perturbations.

We then define the relaxed preservation subspace as:
\begin{equation}
\mathcal{S}
=
\mathrm{span}
\left(
f_b,n_1,\dots,n_k
\right).
\end{equation}
Let
$
A=
\left[
f_b,n_1,\dots,n_k
\right]
$
be the basis matrix of this subspace and the orthogonal projection matrix onto $\mathcal{S}$ is $P_{\mathcal{S}}$. The privacy-related component to be erased is then defined as the orthogonal residual of $f_d$ with respect to the relaxed benign-noise subspace:
\begin{equation}
f_d^{*}
=
\left(I-P_{\mathcal{S}}\right)f_d .
\end{equation}
Therefore, the relaxed optimization objective becomes:
\begin{equation}
\min_{\theta}
\;
\left\|
f_d^{*}
\right\|_2
=
\min_{\theta}
\;
\left\|
\left(I-P_{\mathcal{S}}\right)(f_p-f_m)
\right\|_2 .
\end{equation}
This objective encourages the model to erase the privacy-related differential feature outside the benign-noise subspace, while preserving feature components that are potentially useful for benign OCR ability.

Accordingly, we define the privacy feature erasure loss as:
\begin{equation}
\mathcal{L}_{\mathrm{erase}}
=
\lambda
\left\|
f_d^{*}
\right\|_2
=
\lambda
\left\|
\left(I-P_{\mathcal{S}}\right)(f_p-f_m)
\right\|_2 .
\label{eq:lamda}
\end{equation}
$\lambda$ is an adaptive coefficient used to adjust the strength of privacy feature erasure during back-propagation. It is defined as:
\begin{equation}
\lambda
=
1-
\cos
\left(
f_m+f_d^{*},
f_p
\right),
\end{equation}
where
\begin{equation}
\cos(a,b)
=
\frac{a^{\top}b}{\|a\|_2\|b\|_2}.
\end{equation}
When the orthogonal privacy component $f_d^{*}$ is large, $f_m+f_d^{*}$ is closer to $f_p$, leading to a larger cosine similarity and thus a smaller $\lambda$. This prevents the erasure loss from excessively disrupting the benign feature space. 
Conversely, when $f_d^{*}$ is small, the adaptive coefficient becomes larger, encouraging the model to further erase the remaining privacy-specific information.
In the limiting case where the benign features cannot provide an effective constraint direction for the differential feature, i.e., \(P_{\mathcal{S}}f_d=0\) and thus \(f_d^{*}=f_d\), we have \(f_m+f_d^{*}=f_p\), \(\cos(f_m+f_d^{*}, f_p)=1\) and \(\lambda=0\). This situation arises because the benign subspace \(\mathcal{S}\) is estimated from a single benign image \(I_b\) and its random perturbations, and thus only provides a local approximation of the global benign feature manifold. Consequently, when \(P_{\mathcal{S}}f_d=0\), the known benign subspace imposes no constraint on \(f_d\), and the orthogonal component \(f_d^{*}=f_d\) may contain not only privacy-specific directions but also benign components of other images that lie outside this local approximation. Applying feature-level erasure in this case would force the entire differential feature to vanish and could unintentionally suppress such unseen benign directions. Therefore, we conservatively leave the feature erasure loss inactive and let the output alignment loss dominate privacy suppression.
\(\mathcal{L}_{\mathrm{erase}}\) serves as a auxiliary regularizer that is activated when benign features participate in constraining the differential feature.

Finally, the model is trained with both the output alignment loss and the privacy feature erasure loss:
\begin{equation}
\mathcal{L}
=
\mathcal{L}_{\mathrm{lm}}
+
\mathcal{L}_{\mathrm{erase}},
\end{equation}
where $\mathcal{L}_{\mathrm{lm}}$ denotes the language modeling loss used for output alignment, typically implemented as the cross-entropy loss:
\begin{equation}
\mathcal{L}_{\mathrm{lm}}
=
-
\mathbb{E}_{(I,y)}
\left[
\log p_{\theta}(y\mid I)
\right].
\end{equation}
Through this joint optimization, \textbf{the model is encouraged not only to generate privacy-preserving outputs, but also to suppress privacy-sensitive visual evidence in its internal representations.}

\begin{table}[t]
\centering
\small
\setlength{\tabcolsep}{3.5pt}
\renewcommand{\arraystretch}{1.16}
\definecolor{vipeTableTint}{HTML}{EDF4F8}
\caption{Performance comparison on targeted privacy risk mitigation. We train models on the training set of PRIVACYCARD, and evaluate the privacy leakage rate on the test set. $\Delta$ measures the change in the model's privacy leakage rate after introducing a fixed prefix.}
\label{tab:privacy_leakage}

\begin{adjustbox}{max width=\linewidth}
\begin{tabular}{llcccc}
\toprule
\multirow{2}{*}{Model}
& \multirow{2}{*}{Method}
& \multicolumn{3}{c}{Leakage Rate}\\
\cmidrule(lr){3-5}
&
& w/o prefix
& w/ prefix
& $\Delta$ \\
\midrule
\multirow{6}{*}{LlaVA-onevision1.5}
& --     & 0.7936 & 0.7936      & 0.0000      \\
& SFT           & 0.0000   &  0.8002  &  0.8002  \\
& Memflex      & 0.0004  &    0.8008   &   0.8004    \\
& DINM         & 0.0052   &   0.5734    &   0.5682    \\
& Neural Gate & 0.0000 & 0.7286 & 0.7286 \\
& \cellcolor{vipeTableTint}VIPE (ours) & \cellcolor{vipeTableTint}\textbf{0.0000} & \cellcolor{vipeTableTint}\textbf{0.0004} & \cellcolor{vipeTableTint}\textbf{0.0004}\\

\midrule\midrule

\multirow{5}{*}{Qwen2.5-vl}
& --     & 0.9774 & 0.9774      & 0.0000      \\
& SFT          & 0.0000    & 0.9396   & 0.9396   \\
& Memflex      & 0.2634    & 0.9728      & 0.7094      \\
& DINM         & 0.0006    & 0.9696      & 0.9690       \\
& Neural Gate & 0.0000 & 0.5384 & 0.5384 \\
& \cellcolor{vipeTableTint}VIPE (ours) & \cellcolor{vipeTableTint}\textbf{0.0000} & \cellcolor{vipeTableTint}\textbf{0.0000} & \cellcolor{vipeTableTint}\textbf{0.0000}\\

\bottomrule
\end{tabular}
\end{adjustbox}
\end{table}
\section{Experiments}
\subsection{Settings}
% 我们选择在两种先进的LVLM，qwen2.5-vl-7b和LLaVA-OneVision-1.5-8b，上对算法展开测试。训练阶段，基于Section \ref{privacycard}上提出的PRIVACYCARD，我们在其训练集上使用多种隐私保护算法对基线模型进行微调，使用遗忘集D_f以及保留集D_r分别提高模型的隐私保护能力以及保持模型utility，微调过程中我们使用旋转、拼接等图像增强方法来进一步提高训练数据的多样性。测试阶段，为了充分评估每种算法的效果，我们将评估维度划分为域内安全性、域外安全性、域内有用性以及域外有用性四个大维度。其中，我们使用PRIVACYCARD测试集中包含隐私信息的图像I_p以及良性图像I_b分别测试模型的域内安全性以及域内有用性，并采用Section \ref{}中提到的隐私泄露率以及字错率（CER）两种指标衡量算法对隐私安全性能以及对良性文本提取能力的影响。我们使用Chen等人提出的隐私图像数据集来评估模型域外安全性，该数据集中的图像包含多种粒度、颜色、类型的敏感信息，相比于PRIVACYCARD，该数据集中每张图像包含更多条敏感数据，我们采用隐私泄露率作为评估指标。我们使用OCRBenchv2、MME、POPE评估模型的域外有用性，OCRBenchv2是域外开放视觉问答数据集，MME是域外封闭视觉问答数据集，POPE是评估模型幻觉的数据集，三者从不同维度评估模型域外utility的退化。我们使用ACC评估模型的域外有用性。
We evaluate algorithms on two state-of-the-art LVLMs, Qwen2.5-VL-7B\cite{bai2025qwen25vltechnicalreport} and LLaVA-OneVision-1.5-8B\cite{an2025llavaonevision15fullyopenframework}. During training, we fine-tune the baseline models on the training split of PrivacyCard, introduced in Section~\ref{privacycard}, using various privacy-preserving algorithms. Specifically, we use the forget set $D_f$ to enhance the privacy-preserving capability of the model and the retain set $D_r$ to maintain its utility. In the training process, we apply image augmentation techniques, such as rotation and concatenation, to increase the diversity of the training data.

To comprehensively assess the effectiveness of each algorithm, we divide the evaluation into four major dimensions: in-domain safety, out-of-domain safety, in-domain utility, and out-of-domain utility. For in-domain evaluation, we use the privacy-sensitive images $I_p$ and benign images $I_b$ from the PrivacyCard test set to evaluate in-domain safety and in-domain utility, respectively. We adopt the privacy leakage rate defined in Eq.~\eqref{leakagerate} and the character error rate (CER)\cite{lcvenshtcin1966binary,berg2013unsupervised}, to measure the impact of each algorithm on privacy protection and OCR capability.
For out-of-domain safety, we use the privacy image dataset proposed by Chen et al\cite{chen2025vision}. This dataset contains sensitive information with diverse granularities, colors, and types. Compared with PrivacyCard, each image in this dataset contains a larger number of sensitive data entries. Similar to the evaluation of in-domain safety, we use the privacy leakage rate as the evaluation metric. 
For out-of-domain utility, we evaluate the models on OCRBenchv2\cite{fu2024ocrbenchv2improvedbenchmark}, MME\cite{fu2025mme}, and POPE\cite{li-etal-2023-evaluating}. OCRBenchv2\cite{fu2024ocrbenchv2improvedbenchmark} is an out-of-domain open-ended visual question answering benchmark, MME\cite{fu2025mme} is an out-of-domain closed-ended visual question answering benchmark, and POPE\cite{li-etal-2023-evaluating} is a benchmark for evaluating model hallucination. These three benchmarks assess the degradation of out-of-domain utility from different perspectives. We report accuracy (ACC) to evaluate the out-of-domain utility of the models.

\subsection{Comparison with targeted fine-tuning method}
% 在Section xx中，我们阐述了现有有目标微调隐私保护方法的虚假保护现象，即在前缀条件下模型隐私保护能力失效。本节，我们将给出有目标微调基线方法在前缀增加前后的域内隐私保护能力的量化评估结果。我们评估了四种有目标微调基线方法：SFT使用输出对齐损失微调整个模型，Memflex利用良性样本与敏感样本的梯度定位需要微调的模块，DINM通过微调与隐私最相关的层来实现输出脱敏，Neural Gate将定位从模块粒度转向神经元粒度从而更精细地微调模型。这里，Memflex我们使用[]中的设置以保证不同方法的可比性。表xx中的结果表明，现有有目标微调方法常规输出场景下，即不限制模型输出前缀，能够保持很低的隐私泄露率。然而，当模型的输出前缀被固定，所有基线方法的隐私泄露率均大大提高，增幅达到50%以上。相比之下，VIPE则能够在前缀固定前后均保持很低的隐私泄露率，这说明VIPE能更彻底地消除隐私风险，具有更好的隐私保护鲁棒性。
In Section \ref{phenomenon}, we discuss the phenomenon of illusory protection for existing targeted fine-tuning privacy-preserving methods under prefix-fixing conditions. In this section, we provide quantitative evaluation results for the in-domain privacy-preserving capabilities of targeted fine-tuning baselines before and after prefix addition. We evaluate four targeted fine-tuning baselines: SFT\cite{ouyang2022training} fine-tunes the entire model using an output alignment loss; Memflex\cite{tian2024forget} leverages gradients from benign and sensitive samples to localize modules that require fine-tuning; DINM\cite{wang2024detoxifying} achieves output sanitization by fine-tuning the layers most relevant to privacy; Neural Gate\cite{cao2026neural} shifts the localization from module-level granularity to neuron-level granularity, thereby allowing for more fine-grained model fine-tuning. For Memflex, we adopt the settings in \cite{cao2026neural} to ensure comparability between methods. The results in Table \ref{tab:privacy_leakage} show that existing targeted fine-tuning methods can maintain a very low privacy leakage rate under the conventional output setting, i.e., without constraining the model output prefix. However, when the model output prefix is fixed, the privacy leakage rates of all baseline methods increase substantially, with an increase of over 50\%. In contrast, VIPE maintains a very low privacy leakage rate both before and after fixing the prefix, indicating that VIPE can more thoroughly eliminate privacy risks and exhibits better robustness in privacy protection.

\subsection{Comparison with visual privacy erasure baseline methods}
% 根据Section \ref{}的结果，我们了解到现有有目标隐私加固方法在隐私保护鲁棒性上存在缺陷，即当模型输出前缀改变时算法隐私保护能力失效。通过Section \ref{}的实验，我们将该现象归结于模型内部视觉隐私信息在有目标隐私加固后仍然存在。由于目前很少有研究关注视觉特征维度的隐私信息擦除，为了保证结果对比的公平性并突出我们方法的优越性，我们引入了多种基线视觉隐私信息擦除方法的效果作为对比。基线方法包括直接利用MSE对齐敏感图像I_p以及对应掩码图像I_m的视觉特征，使用余弦相似度对齐I_p与I_m的方向，以及使用kl散度对齐I_p与I_m的分布。此外，在隐私安全性评估中，我们比较了模型在输出前缀确定前后隐私泄露率的差异，以此推断模型内部隐私视觉特征擦除的情况，结果见表xx所示。
VIPE seeks to erase visual privacy information by aligning the features of sensitive images with those of their masked counterparts while preserving benign visual features. A natural question is whether directly aligning these features with standard objectives can already achieve robust privacy protection without compromising visual utility. We therefore compare VIPE with visual privacy erasure baselines. This comparison evaluates whether benign-subspace constraint of VIPE offers a better privacy–utility trade-off than direct feature alignment.

\begin{figure*}[t]
    \centering
    \includegraphics[width=1.\linewidth]{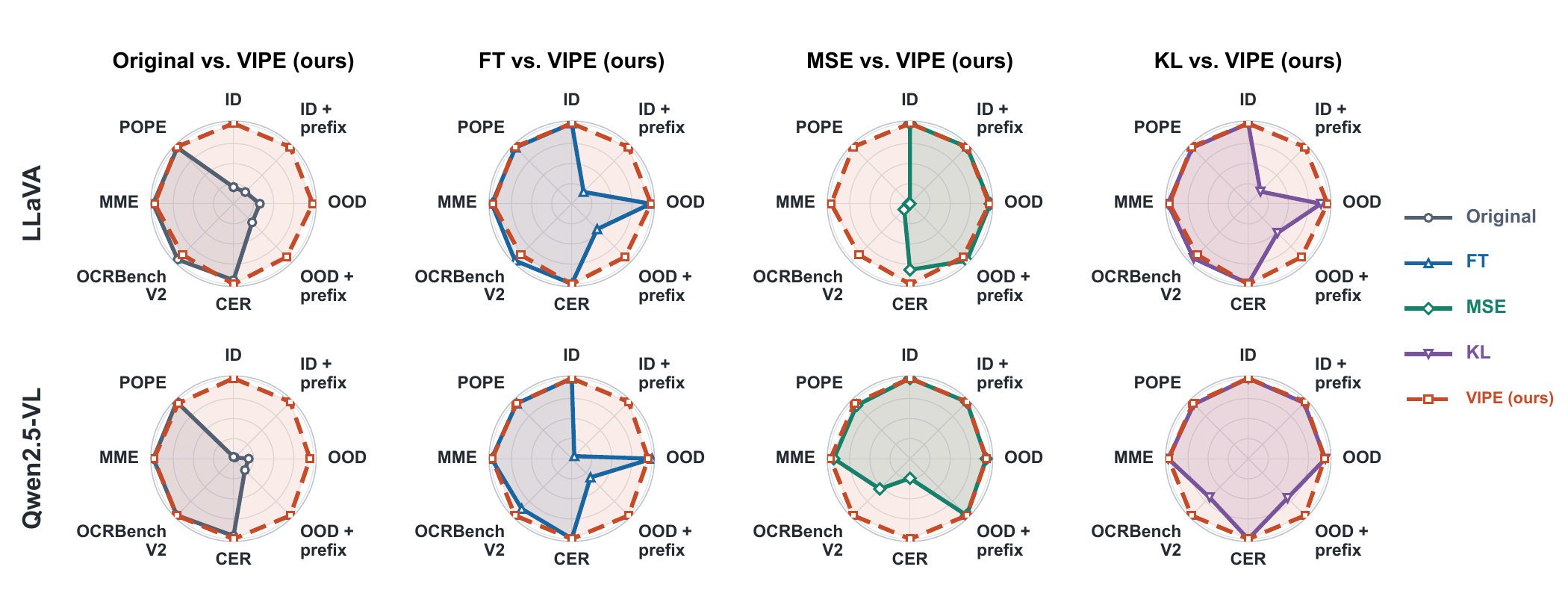}
    \caption{Performance of each visual privacy erasure on Qwen2.5-vl and LLaVA-Onevision-1.5. For the privacy leakage rate and CER, we use one minus the original value as the radar-chart metric, so that larger values indicate better performance across all dimensions. The maximum value of each dimension is set to the maximum value among all compared methods for the corresponding model along that dimension. ID denotes in-domain privacy risk, OOD denotes out-of-domain privacy risk, and prefix indicates that the output prefix is fixed.}
    \label{fig:radar_erase}
\end{figure*}

We construct two baselines: directly aligning the visual features of $I_p$ and $I_m$ using MSE, 
% aligning the feature directions of $I_p$ and $I_m$ using cosine similarity, 
and aligning their feature distributions using KL divergence\cite{kullback1951information}. 
In the safety evaluation, we follow the prefix-injection setting in Section \ref{phenomenon} and compare privacy leakage rates before and after injecting the model’s output prefix. This comparison assesses whether each method effectively erases visual privacy feature and provides a better measure of privacy protection robustness.
The results are shown in Fig. \ref{fig:radar_erase}.

% 根据表xx，我们发现直接使用mse对齐I_p与I_m的特征可能严重损害模型性能，llava经过微调后几乎丧失了域外视觉问答的能力，而qwen2.5经过微调后开放问答与域外封闭问答性能大幅降低。利用余弦相似度对齐I_p与I_m的隐私特征擦除方法虽然在性能上损失要小于基于mse的方法，但其在域外开放问答上的表现仍出现严重下滑，降幅接近30%。而不引入视觉隐私特征擦出的微调方法以及基于kl的隐私特征擦除方法仍然存在隐私保护能力不鲁棒的问题，模型在前缀固定的情境下隐私保护能力极大降低。相比之下，我们的方法能够在各种情境下取得较好的效果，不仅能够具备较鲁棒的隐私保护能力，还能几乎不损失模型的可用性。在图xx中，我们可以进一步观察到我们方法的优势，每个雷达图显示了我们方法与其他方法在所有维度微调效果的比较。我们的方法在每个评估维度均优于或comparable与其他所有方法。

As shown in Fig.~\ref{fig:radar_erase}, directly aligning the features of $I_p$ and $I_m$ using MSE can severely degrade model utility. After fine-tuning with MSE, LLaVA\cite{an2025llavaonevision15fullyopenframework} almost loses its ability to perform out-of-domain visual question answering, while Qwen2.5\cite{bai2025qwen25vltechnicalreport} shows a substantial decline in open-ended question-answering performance on OCRBench v2\cite{fu2024ocrbenchv2improvedbenchmark}.
% and out-of-domain closed-ended question answering (MME\cite{fu2025mme} and POPE\cite{li-etal-2023-evaluating}). 
% Although the privacy feature erasure method based on cosine similarity incurs a smaller performance loss than the MSE-based method, it still leads to severe degradation in out-of-domain open-ended visual question answering, with a relative performance drop of nearly 30\% compared with the original accuracy.
Privacy protection remains insufficiently robust for both fine-tuning methods that do not explicitly erase visual privacy features and KL-based privacy feature erasure methods. Specifically, when the output prefix is fixed, the privacy-preserving capability of these methods decreases dramatically. 
In contrast, VIPE performs well across all evaluation dimensions. VIPE not only provides robust privacy protection but also preserves the usability of the model.
% The advantage of VIPE can be further observed in Fig. \ref{fig:radar_erase}, where each radar chart compares the performance of our method with those of other methods across all evaluation dimensions. Our method outperforms or remains comparable to all methods on every evaluation dimension.

\subsection{Comparison with untargeted forgetting method}
\label{sec:forget}
% Required preamble packages:
% \usepackage{booktabs}
% \usepackage{multirow}
% \usepackage[table]{xcolor} % If xcolor is already loaded, use \usepackage{colortbl}.
% \usepackage{adjustbox}
% Insert using \input{untargeted_method_v3.tex}.
% All supplied observations are preserved; only decimal formatting is unified.
% Ranking excludes Original and uses distinct values; ties receive the same mark.
\begin{table*}[t]
\centering
\begingroup
\definecolor{vipeTableTint}{HTML}{EDF4F8}
\definecolor{vipeTableAccent}{HTML}{245675}
\small
\setlength{\tabcolsep}{3.5pt}
\renewcommand{\arraystretch}{1.16}
\caption{Safety-utility comparison of untargeted unlearning methods.
Within each model, \textbf{bold} and \underline{underlined} values denote the best and second-best results among the four methods, respectively.}
\label{tab:untargeted_comparison}
\begin{adjustbox}{max width=\textwidth}
\begin{tabular}{@{}llcccccccc@{}}
\toprule
\multirow{2}{*}{Model}
& \multirow{2}{*}{Method}
& \multicolumn{2}{c}{In-domain Safety $\downarrow$}
& \multicolumn{2}{c}{Out-of-domain Safety $\downarrow$}
& In-domain Utility $\downarrow$
& \multicolumn{3}{c}{Out-of-domain Utility $\uparrow$} \\
\cmidrule(lr){3-4}
\cmidrule(lr){5-6}
\cmidrule(lr){7-7}
\cmidrule(lr){8-10}
& & w/o prefix & w/ prefix & w/o prefix & w/ prefix & CER & OCRBenchv2 & MME & POPE \\
\midrule
& Original & 0.7936 & 0.7936 & 0.6726 & 0.6726 & 0.2108 & 0.329 & 0.8736 & 0.8966 \\
\cmidrule(lr){2-10}
\morecmidrules
\cmidrule(lr){2-10}
& GA & \textbf{0.0000} & \textbf{0.0000} & \underline{0.0044} & \underline{0.0059} & 0.3829 & 0.125 & \textbf{0.8719} & 0.8780 \\
& NPO & \textbf{0.0000} & \textbf{0.0000} & 0.0278 & 0.0281 & 5.0131 & \underline{0.279} & 0.8500 & \underline{0.9036} \\
& MMUnlearner & \textbf{0.0000} & \textbf{0.0000} & \textbf{0.0018} & \textbf{0.0018} & \underline{0.1940} & 0.104 & 0.8496 & 0.9000 \\
\multirow{-5}{*}{LLaVA} & \cellcolor{vipeTableTint}VIPE (ours) & \cellcolor{vipeTableTint}\textbf{0.0000} & \cellcolor{vipeTableTint}\underline{0.0004} & \cellcolor{vipeTableTint}0.0146 & \cellcolor{vipeTableTint}0.0648 & \cellcolor{vipeTableTint}\textbf{0.1698} & \cellcolor{vipeTableTint}\textbf{0.301} & \cellcolor{vipeTableTint}\underline{0.8572} & \cellcolor{vipeTableTint}\textbf{0.9070} \\

\addlinespace[3pt]
\midrule
& Original & 0.9774 & 0.9774 & 0.8165 & 0.8165 & 0.0560 & 0.328 & 0.8698 & 0.8890 \\
\cmidrule(lr){2-10}
\morecmidrules
\cmidrule(lr){2-10}
& GA & \textbf{0.0000} & \textbf{0.0000} & 0.4346 & 0.4481 & 0.0501 & 0.100 & 0.7700 & 0.8223 \\
& NPO & \textbf{0.0000} & \textbf{0.0000} & \textbf{0.0000} & \underline{0.0165} & 1.0000 & \underline{0.220} & 0.8504 & \textbf{0.8896} \\
& MMUnlearner & \textbf{0.0000} & \textbf{0.0000} & \underline{0.0007} & \textbf{0.0011} & \textbf{0.0169} & 0.205 & \underline{0.8622} & 0.8593 \\
\multirow{-5}{*}{Qwen2.5} & \cellcolor{vipeTableTint}VIPE (ours) & \cellcolor{vipeTableTint}\textbf{0.0000} & \cellcolor{vipeTableTint}\textbf{0.0000} & \cellcolor{vipeTableTint}0.0746 & \cellcolor{vipeTableTint}0.0867 & \cellcolor{vipeTableTint}\underline{0.0180} & \cellcolor{vipeTableTint}\textbf{0.329} & \cellcolor{vipeTableTint}\textbf{0.8639} & \cellcolor{vipeTableTint}\underline{0.8660} \\
\bottomrule
\end{tabular}
\end{adjustbox}
\endgroup
\end{table*}

% 无目标遗忘方法通常通过梯度上升或负偏好优化（NPO）的方法让模型对敏感问题的输出远离原始输出。自回归语言模型将输出序列的概率分解为一系列条件 token 概率的乘积，其中每个 token 都基于其前序上下文生成。在这一建模方式下，对隐私敏感目标执行梯度上升会增大后续敏感 token 的负对数似然损失，从而抑制其条件生成概率。这种算法天然提供了对隐私前缀续写的防御能力，因为每个正确敏感信息的token基于前序上下文的损失都被放大、生成概率都被降低，模型本质上失去了生成连续敏感信息的能力。然而，这些方法通常因为损失函数无界、优化方向不稳定等原因导致模型性能退化。表xx中对比了我们的方法与多种无目标遗忘算法的性能。
Untargeted unlearning methods typically use gradient ascent (GA)\cite{thudi2022unrolling,huo2025mmunlearner} or Negative Preference Optimization (NPO)\cite{zhang2024negative} to move the model’s responses to sensitive queries away from the original outputs. Autoregressive LVLMs factorize the probability of an output sequence into a product of token-level probabilities, where each token is predicted conditioned on the preceding context\cite{cover1991elements,radford2018improving,liu2023visual}. Under this formulation, applying GA to privacy-sensitive targets increases the negative log-likelihood of subsequent sensitive tokens. 
Such algorithms naturally provide a defense against privacy-prefix continuation. For each token in the sensitive information, the loss conditioned on the preceding context is amplified, and its generation probability is reduced. As a result, the model largely loses the ability to generate consecutive sensitive information. However, these methods often lead to performance degradation. This may be due to issues such as unbounded loss functions and unstable optimization directions\cite{huang2026gradient}. Table \ref{tab:untargeted_comparison} reports the performance of our method and several untargeted unlearning algorithms.
% 我们发现无目标遗忘方法能够使模型具备良好的隐私保护能力，即使输出前缀固定，这些方法仍能具备鲁棒的安全性，这与预期一致。然而，这些无目标微调的方法由于破坏了模型连续输出某些信息的能力，这使得模型开放问答（OCRBench v2）情景下的表现出现严重退化，对于输出长度较短的封闭式问答 (MME and POPE)，现有无目标遗忘方法性能退化程度有限。这种结果可能表明无目标微调方法使得模型无法输出某些信息，但是模型内部仍然能够感知这些信息。相比于这些无目标遗忘方法，我们的方法对模型性能影响更小，无论是在开放问答还是封闭问答情景下。
Our results show that untargeted unlearning methods endow models with strong privacy-preserving capabilities. Even when the output prefix is fixed, these methods still maintain robust safety. 
However, since untargeted unlearning disrupts the model’s ability to continuously generate certain information, it leads to severe degradation in open-ended question-answering performance on OCRBench v2\cite{fu2024ocrbenchv2improvedbenchmark}.
In contrast, MME\cite{fu2025mme} and POPE\cite{li-etal-2023-evaluating} involve closed-ended questions with shorter answers. Existing untargeted unlearning methods lead to only limited performance degradation on these tasks.
These results may suggest that untargeted unlearning prevents the model from correctly answering benign queries, while the model may still know the answers.
Compared with these untargeted unlearning methods, VIPE causes substantially less utility degradation, in both open‑ended and closed‑ended QA settings.

% 此外，我们还评估了I_p对微调后模型回答良性问题的影响程度。具体来说，在输入图像为I_p、输入文本问题为良性问题Q_b（数学计算、常识问答等）的情况下，评估模型回答良性问题Q_b的能力是否会退化。我们使用崩溃率来量化模型对敏感图像I_p的敏感性，崩溃率表示所有测试样本中输出崩溃的样本占比，模型在样本<I_p，Q_b>输入下输出崩溃当且仅当模型输出内容与Q_b不相关。如图xx所示，对所有无目标遗忘方法而言，I_p都会让模型失去回答良性问题的能力。相比之下，我们的方法能够让模型在输入图像为I_p的情况下仍能够正常回答其他良性问题Q_b，在模型性能维持上表现更优。
\begin{figure}[t]
    \centering
    \includegraphics[width=\linewidth]{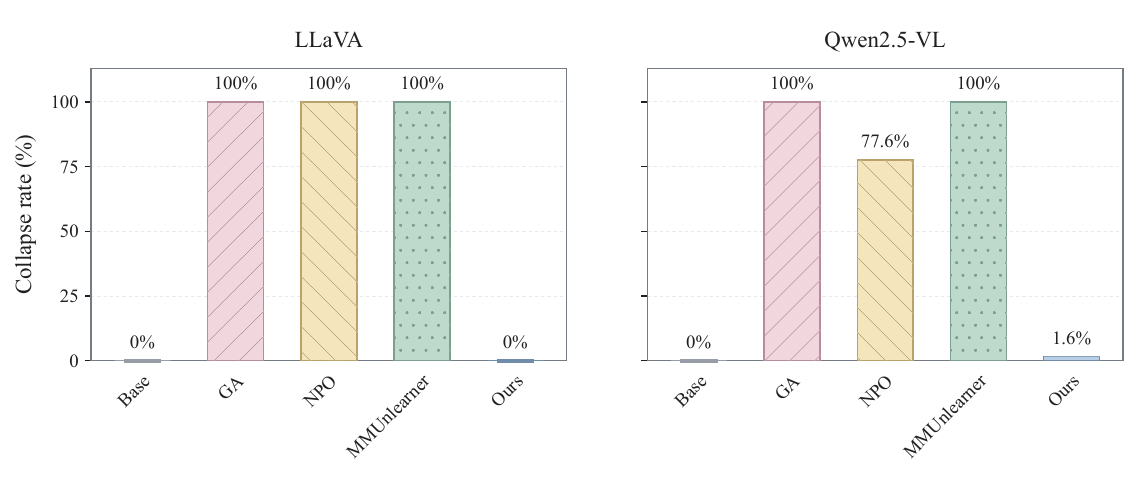}
    \caption{Collapse rates of LLaVA and Qwen2.5 fine-tuned with different privacy-preserving methods on sensitive image–benign question pairs. The collapse rate is measured as the proportion of responses irrelevant to the benign question.}
    \label{fig:collapse_rate}
\end{figure}
In addition, we evaluate whether the fine-tuned model can still answer benign questions when the sensitive image $I_p$ is provided as input. In this evaluation, the image input is fixed as $I_p$, while the text input is a benign question $Q_b$, such as a mathematical calculation or a common sense question. We then examine whether the model’s ability to answer $Q_b$ is degraded in the presence of $I_p$.
To quantify how strongly the sensitive image $I_p$ interferes with the model’s ability to answer benign questions, we introduce the collapse rate, defined as the proportion of test samples that result in collapsed outputs. For an input sample $\langle I_p, Q_b \rangle$, we regard the output as collapsed if it is irrelevant to the benign question $Q_b$.
As shown in Fig. \ref{fig:collapse_rate}, all untargeted unlearning methods suffer from a high collapse rate. When $I_p$ is given as the image input, the models often produce responses that are not related to $Q_b$. In contrast, VIPE maintains a low collapse rate, further demonstrating that VIPE better preserves the model’s utility.

\subsection{Ablation experiments}
\begin{table*}[t]
\centering
\scriptsize
\definecolor{vipeTableTint}{HTML}{EDF4F8}
\small
\setlength{\tabcolsep}{3.5pt}
\renewcommand{\arraystretch}{1.16}
\caption{Ablation study of privacy erasure loss and conditional relaxation. 
$\mathcal{L}_{\mathrm{lm}}$ denotes the standard language modeling loss, and 
$\mathcal{L}_{\mathrm{erase}}$ denotes the privacy erasure loss. 
$k$ denotes the number of noise vectors used for conditional relaxation.  
}
\label{tab:ablation_noise}
% \begin{adjustbox}{width=\textwidth}
\begin{adjustbox}{max width=\textwidth}
% \begin{tabular}{@{}llcccccccc@{}}
\begin{tabular}{lllcccccccc}
\toprule
\multirow{2}{*}{Backbone}
& \multirow{2}{*}{Loss}
& \multirow{2}{*}{$k$}
& \multicolumn{2}{c}{In-domain Safety $\downarrow$}
& \multicolumn{2}{c}{Out-of-domain Safety $\downarrow$}
& In-domain Utility $\downarrow$
& \multicolumn{3}{c}{Out-of-domain Utility $\uparrow$}\\
\cmidrule(lr){4-5}
\cmidrule(lr){6-7}
\cmidrule(lr){8-8}
\cmidrule(lr){9-11}
& & & w/o prefix & w/ prefix & w/o prefix & w/ prefix & CER & OCRBenchv2 & MME & POPE\\
\midrule

\multirow{6}{*}{LLaVA}
& --
& --
& 0.7936
& 0.7936
& 0.6726
& 0.6726
& 0.2108
& 0.329
& 0.8736
& 0.8966\\
& $\mathcal{L}_{\mathrm{lm}}$
& --
& 0.0000
& 0.7912
& 0.0205
& 0.5536
& 0.1729
& 0.335
& 0.8689
& 0.8980\\
& $\mathcal{L}_{\mathrm{lm}} + \mathcal{L}_{\mathrm{erase}}$
& 0
& 0.0000
& 0.0146
& 0.0025
& 0.1731
& 0.1682
& 0.288
& 0.8689
& 0.9113\\
& $\mathcal{L}_{\mathrm{lm}} + \mathcal{L}_{\mathrm{erase}}$
& 1
& 0.0000
& 0.0038
& 0.0120
& 0.1431
& 0.1728
& 0.304
& 0.8668
& 0.9053\\
& \cellcolor{vipeTableTint}$\mathcal{L}_{\mathrm{lm}} + \mathcal{L}_{\mathrm{erase}}$
& \cellcolor{vipeTableTint}2
& \cellcolor{vipeTableTint}0.0000
& \cellcolor{vipeTableTint}0.0004
& \cellcolor{vipeTableTint}0.0146
& \cellcolor{vipeTableTint}0.0648
& \cellcolor{vipeTableTint}0.1698
& \cellcolor{vipeTableTint}0.301
& \cellcolor{vipeTableTint}0.8572
& \cellcolor{vipeTableTint}0.9070\\
& $\mathcal{L}_{\mathrm{lm}} + \mathcal{L}_{\mathrm{erase}}$
& 3
& 0.0000
& 0.0004
& 0.0472
& 0.0596
& 0.1768
& 0.296
& 0.8593
& 0.9063\\

\midrule

\multirow{6}{*}{Qwen2.5}
& --
& --
&  0.9774
& 0.9774
& 0.8165
& 0.8165
& 0.0560
& 0.328
& 0.8698
& 0.8890\\
& $\mathcal{L}_{\mathrm{lm}}$
& --
& 0.0000
& 0.9546
& 0.0285
& 0.6993
& 0.0219
& 0.290
& 0.8719
& 0.8620\\
& $\mathcal{L}_{\mathrm{lm}} + \mathcal{L}_{\mathrm{erase}}$
& 0
& 0.0000
& 0.0002
& 0.0563
& 0.1047
& 0.0767
& 0.294
& 0.8567
& 0.8616\\
& $\mathcal{L}_{\mathrm{lm}} + \mathcal{L}_{\mathrm{erase}}$
& 1
& 0.0000
& 0.0000
& 0.0296
& 0.0432
& 0.0209
& 0.290
& 0.8213
& 0.8486\\
& \cellcolor{vipeTableTint}$\mathcal{L}_{\mathrm{lm}} + \mathcal{L}_{\mathrm{erase}}$
& \cellcolor{vipeTableTint}2
& \cellcolor{vipeTableTint}0.0000
& \cellcolor{vipeTableTint}0.0000
& \cellcolor{vipeTableTint}0.0746
& \cellcolor{vipeTableTint}0.0867
& \cellcolor{vipeTableTint}0.0180
& \cellcolor{vipeTableTint}0.329
& \cellcolor{vipeTableTint}0.8639
& \cellcolor{vipeTableTint}0.8660\\
& $\mathcal{L}_{\mathrm{lm}} + \mathcal{L}_{\mathrm{erase}}$
& 3
& 0.0000
& 0.0012
& 0.1149
& 0.2138
& 0.0338
& 0.319
& 0.8702
& 0.8646\\

\bottomrule
\end{tabular}
\end{adjustbox}
\end{table*}
% 在本节中，我们对隐私擦除损失 $\mathcal{L}_{\mathrm{erase}}$ 中的条件松弛参数 $k$ 进行消融分析，并考察其对隐私保护与视觉效用的影响。如表~\ref{tab:ablation_noise} 所示，在 LLaVA 上，仅使用 $\mathcal{L}_{\mathrm{lm}}$ 已能将无前缀条件下的域内外隐私泄露率降至零附近，但带前缀条件下仍有较高泄露率。引入 $\mathcal{L}_{\mathrm{erase}}$ 后，带前缀域外安全显著改善。当 $k$ 从 0 增加到 3 时，隐私泄露率由 0.1731 下降到 0.0596，说明适度放松约束有助于稳定擦除过程并降低前缀触发下的泄露风险。不过，$k$ 过大也会削弱擦除效果的稳定性，尤其在无前缀域外安全上，指标从 0.0025 上升到 0.0472，表明保留空间扩大后，更多隐私信息可能残留。效用方面，$k=2$ 在 LLaVA 上取得了较均衡的结果：域内良性文本识别CER为 0.1698，POPE上的准确率取得为0.9070，同时带前缀域外安全保持在 0.0648，表现出较好的安全-效用折中。
% 在 Qwen2.5 上，单独使用 $\mathcal{L}_{\mathrm{lm}}$ 同样无法充分抑制带前缀场景下的隐私泄露，域内外隐私泄露率极高。加入 $\mathcal{L}_{\mathrm{erase}}$ 后，$k=1$ 在安全性上最强，带前缀域外安全降至 0.0432，但其效用明显受损，MME上的准确率下降到 0.8213。相比之下，$k=2$ 在 Qwen2.5 上取得了更稳妥的平衡。域内效用指标以及OCRBench v2上的指标也达到最佳，同时带前缀域外隐私泄露率保持在0.0867。虽然 $k=3$ 在 MME上略有优势，但其带前缀域外隐私泄露率明显恶化到 0.2138，说明过大的松弛会削弱隐私擦除的有效性。
% 综合来看，$\mathcal{L}_{\mathrm{erase}}$ 对隐私保护至关重要，而 $k$ 控制了擦除约束的松弛程度：过小会限制模型保留必要的视觉信息，过大则会留下更多隐私残留。结合两个 backbone 上的安全性与效用表现，我们最终选择 $K=2$ 作为主实验设置。
First, we conduct an ablation study on the conditional relaxation parameter $k$ in the privacy erasure loss $\mathcal{L}_{\mathrm{erase}}$ and examine its impact on privacy protection and visual utility. 

As shown in Table~\ref{tab:ablation_noise}, on LLaVA\cite{an2025llavaonevision15fullyopenframework}, optimizing only with $\mathcal{L}_{\mathrm{lm}}$ is already sufficient to reduce the in-domain and out-of-domain privacy leakage rates to nearly zero, but substantial leakage still remains under the prefix-fixing setting. After introducing $\mathcal{L}_{\mathrm{erase}}$, the out-of-domain safety under the prefix-fixing setting improves markedly. As $k$ increases from 0 to 3, the privacy leakage rate decreases from 17.31\% to 5.96\%, indicating that a moderate relaxation of the restriction helps stabilize the erasure process and mitigate prefix-triggered leakage. 
However, an overly large $k$ reduces privacy protection under the conventional setting without prefix fixing. In this setting, the out-of-domain privacy leakage rate increases from 0.25\% to 4.72\%, suggesting that a larger preservation subspace may retain more privacy-related information. 
In terms of utility, $k=2$ achieves a favorable balance on LLaVA\cite{an2025llavaonevision15fullyopenframework}, yielding a CER of 0.1698 in the in-domain utiliy evaluation and an accuracy of 90.7\% on POPE, while maintaining an out-of-domain privacy leakage rate of 6.48\% under the prefix-fixing setting, demonstrating a strong trade-off between safety and utility.

For Qwen2.5\cite{bai2025qwen25vltechnicalreport}, using only $\mathcal{L}_{\mathrm{lm}}$ is also insufficient to suppress privacy leakage under the prefix-fixing setting, with both in-domain and out-of-domain leakage rates remaining very high. After introducing $\mathcal{L}_{\mathrm{erase}}$, $k=1$ achieves the strongest privacy protection, reducing the out-of-domain privacy leakage rate under the prefix-fixing setting to 4.32\%. However, the utility of the model drops clearly, with MME accuracy decreasing to 82.13\%. By contrast, $k=2$ provides a more balanced trade-off on Qwen2.5\cite{bai2025qwen25vltechnicalreport}. It achieves the best in-domain utility and the best performance on OCRBench v2\cite{fu2024ocrbenchv2improvedbenchmark}, while keeping the out-of-domain privacy leakage rate under the prefix-fixing setting at 8.67\%. Although $k=3$ yields a slight advantage on MME, its out-of-domain privacy leakage rate under the prefix-fixing setting deteriorates substantially to 21.38\%, indicating that excessive relaxation weakens the effectiveness of privacy erasure.

Overall, $\mathcal{L}_{\mathrm{erase}}$ is crucial for privacy protection, while $k$ controls the degree of relaxation in the erasure constraint. Considering the safety and utility performance in the two models, we ultimately choose $k=2$ as the main experimental setting.
\begin{figure}[h]
    \centering
    \includegraphics[width=\linewidth]{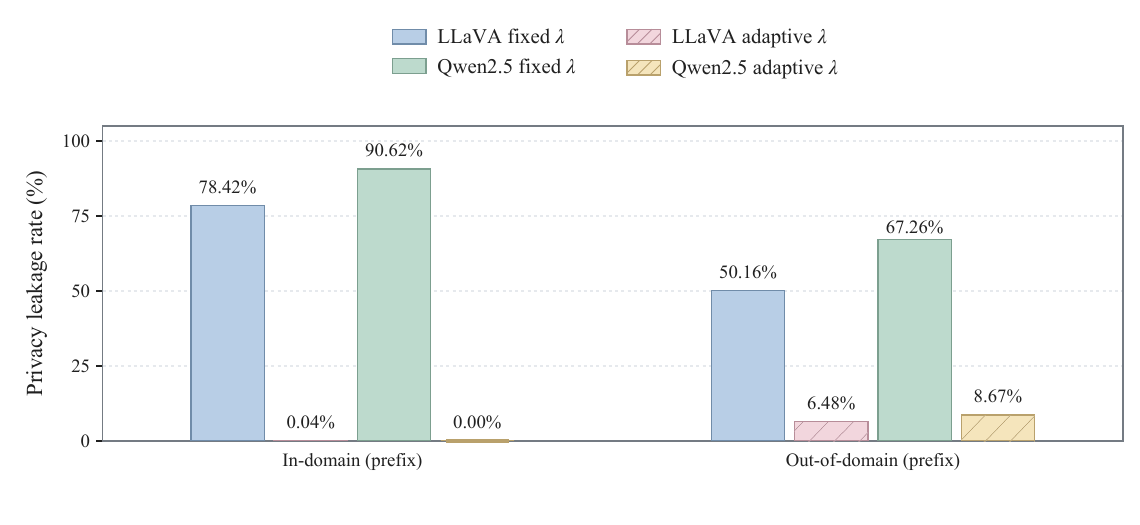}
    \caption{Privacy leakage rate of LLaVA and Qwen2.5. Effect of $\lambda$'s adaptivity on privacy preservation under the prefix-fixing setting, including in-domain and out-of-domain tests.
}
    \label{fig:privacy_leakage_bar}
\end{figure}

Additionally, we conduct an ablation study to examine the effect of the adaptive $\lambda$ in Eq~\eqref{eq:lamda}. 
% 具体而言，我们评估了在the adaptive $\lambda$取固定值时算法的表现，该固定值取the adaptive $\lambda$在所有训练样本中的均值，以此评估$\lambda$的自适应性是否真正提高了算法表现。根据图xx的结果，我们发现当$\lambda$的值固定后，模型在前缀条件下的隐私泄露率大大提高，这体现了$\lambda$自适应性的重要性。
Specifically, we compare the adaptive $\lambda$ with a fixed one which is held constant at its mean value over all training samples.
The mean value is chosen deliberately because it preserves the average strength of the privacy-erasure loss while removing only the sample-wise variation in $\lambda$. 
This provides a controlled test of adaptivity: if the benefit of adaptive $\lambda$ is merely due to the erasure strength, the fixed $\lambda$ would perform similarly. 
According to the results in Fig.~\ref{fig:privacy_leakage_bar}, we find that for both Qwen2.5\cite{bai2025qwen25vltechnicalreport} and LLaVA\cite{an2025llavaonevision15fullyopenframework}, once $\lambda$ is fixed, the privacy leakage rate of the model under prefix conditions increases substantially, demonstrating that the adaptivity of $\lambda$ is crucial for privacy protection.

\subsection{Performance on Mixed-Content Images}
\begin{table}[t]
\centering
\caption{Performance on mixed-content images. Leakage Rate measures the privacy leakage rate, and CER denotes the character error rate of extracted benign text.}
\label{tab:mix_preformance}

\begin{adjustbox}{max width=\linewidth,keepaspectratio}
\begin{tabular}{llcc}
\toprule
Model
& Method
& Leakage Rate $\downarrow$
& CER $\downarrow$\\
\midrule

\multirow{2}{*}{LLaVA}
& Baseline     & 0.8186 & 0.7023  \\
& VIPE & 0.0012 & 0.1640 \\
\midrule\midrule
\multirow{2}{*}{Qwen2.5}
& Baseline     & 0.9776 & 0.5866       \\
& VIPE & 0.0000 & 0.0236\\

\bottomrule
\end{tabular}
\end{adjustbox}
\end{table}
% 混合内容图像中的表现
% 现实情况下，模型接收的图像通常为混合内容，即同时包含敏感信息与非敏感信息。为了评估算法在这类图像中的表现，我们通过向PrivacyCard中的良性图像注入敏感信息来构造混合内容图像，并让模型提取该类图像中的文本信息。我们通过隐私泄露率以及字符错误率两种指标分别评估模型的安全性以及可用性，结果展示在表格xx中。结果显示，VIPE能够在提取混合内容图像的文本信息时正确过滤敏感信息，极大降低了隐私泄露率。同时，模型能够极大程度上保留对良性文本信息的提取能力。这种结果进一步证明了VIPE在现实场景的应用价值。
In realistic settings, the images that models process are typically mixed-content, containing both sensitive and non-sensitive information. To assess the algorithm's performance on such images, we construct mixed-content images by injecting sensitive information into benign images from the test set of PrivacyCard and then prompt the model to extract textual information. We adopt two metrics, the privacy leakage rate and the character error rate (CER), to evaluate security and utility, respectively. 
We calculate CER by comparing the complete output of the model with the benign text content. This implies that when the model leaks private information, its output deviates further from the benign text, leading to a higher CER. As a result, CER values are generally higher for mixed-content images.
The results are presented in Table~\ref{tab:mix_preformance}. We find that VIPE can filter out sensitive information during text extraction from mixed-content images, significantly reducing the privacy leakage rate, while largely maintaining the ability to extract benign textual information. This further demonstrates the practical value of VIPE in real-world applications.

\subsection{Stability Analysis}
\begin{figure*}[t]
    \centering
    \includegraphics[width=\linewidth]{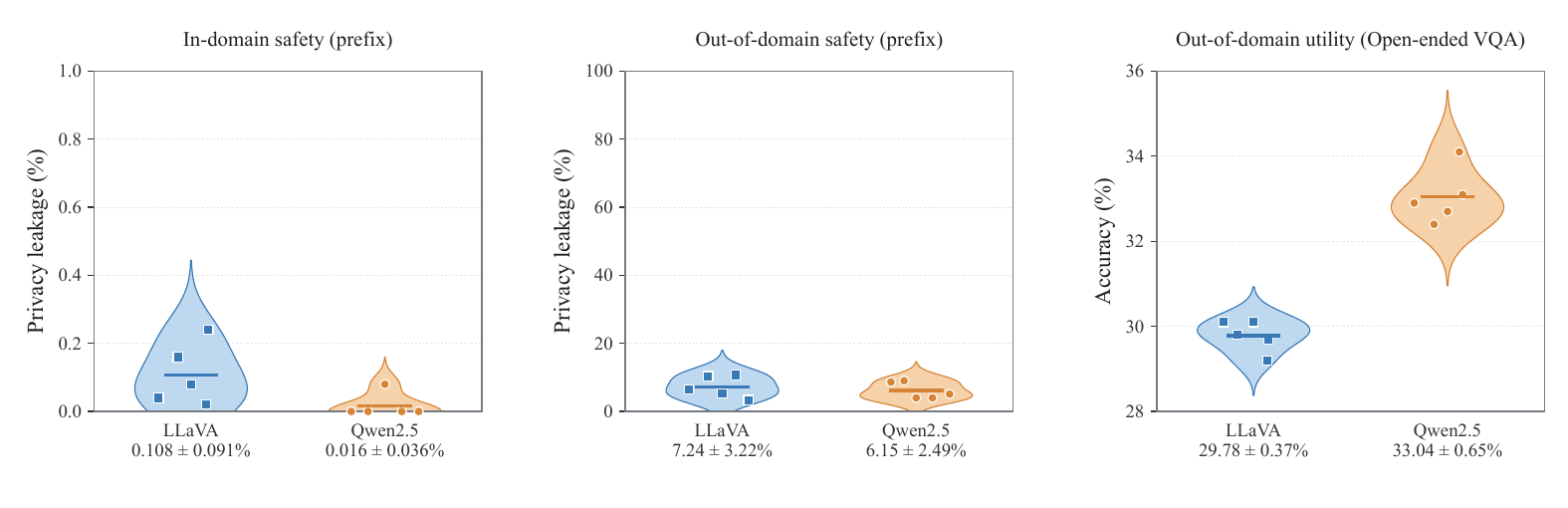}
    \caption{Stability Analysis for VIPE. Performance of VIPE under the prefix-fixing setting on in-domain and out-of-domain safety evaluations and out-of-domain open-ended question datasets. We repeat VIPE five times across different models and quantify the variability of the algorithm in terms of safety and utility.
    }
    \label{fig:stability}
\end{figure*}
% 鉴于VIPE引入随机噪声作为条件松弛组件，我们需要分析这种随机性对算法的影响。为此，我们重复开展5次实验，评估模型在前缀注入下的域内安全性与域外安全性，以及模型在域外开放问答场景下的性能。这三个指标具有很强代表性，前缀条件下的安全性指标相比于不加前缀的情况更能够反映隐私擦除的效果，而开放问答性能指标相比于封闭问答指标难度更大，而且根据Section \ref{sec:forget}的结果，该指标更容易受到模型特征空间变化的影响，即指标更敏感。重复实验结果如图xx所示，我们发现随机噪声在一定程度上对VIPE的效果产生波动，这种波动在域内安全性以及开放问答性能两方面较小，而对域外安全性影响更大。但即使域外安全性的结果存在波动，对于LLaVA以及Qwen2.5两个模型而言，算法依旧能够将隐私泄露率控制在相对低的水平，均值控制在7\%左右。这表明VIPE的结果有较好的可复现性以及稳定性，噪声的随机性并不会主导算法最终的效果。
Since VIPE introduces random noise as a conditional relaxation module, the stochasticity induced by this design may affect both the algorithm's performance and the privacy-erasure behavior. Therefore, it is necessary to evaluate the stability of VIPE with respect to this source of randomness. To this end, we conduct five independent runs under identical experimental configurations. We examine the model behavior along three representative dimensions: in-domain safety under prefix injection, out-of-domain safety under prefix injection, and model performance on out-of-domain open-ended question answering.
These metrics were selected for the following reasons. First, safety under prefix-fixing setting provides a more direct assessment of the privacy leakage risk after privacy erasure. Second, open-ended question answering imposes substantially greater demands on the model's knowledge expression and generative ability than closed-ended question answering. 
% Moreover, as demonstrated in Section \ref{sec:forget}, this metric is more susceptible.
The results of the repeated runs are presented in Fig. \ref{fig:stability}. Overall, injected random noise induces a certain degree of run-to-run variation in the performance of VIPE. The magnitude of this variation differs across evaluation dimensions. In particular, the variations in in-domain safety and out-of-domain open-ended question-answering performance are relatively limited, whereas out-of-domain safety exhibits noticeably greater variability. Despite this variation, VIPE consistently maintains the privacy leakage rate at a relatively low level throughout all repeated runs for both LLaVA\cite{an2025llavaonevision15fullyopenframework} and Qwen2.5\cite{bai2025qwen25vltechnicalreport}, with the mean leakage rate remaining approximately $7\%$. These results indicate that, although the stochasticity introduced during conditional relaxation leads to a measurable degree of uncertainty, its effect remains well controlled and does not constitute the primary determinant of the algorithm's overall performance. The results demonstrate that VIPE exhibits strong consistency, reproducibility, and stability across independent runs with different noise realizations.

\section{Conclusion}
% 本文中我们揭漏了当前针对LVLM的有目标隐私保护方法在感知任务上存在的虚假保护现象，即这类隐私保护方法会在输出前缀固定的情况下失效，并对该现象的起因进行了进一步探究。我们发现这种现象可能源于微调前后模型内部视觉图像特征保持高度一致性，这种一致性使得模型无法在常规情况下输出敏感信息，但仍能正确感知图像中的敏感信息，也就是说这类方法无法本质上消除隐私安全风险。为了能够更本质的消除隐私风险，我们提出了VIPE，一种基于隐私视觉特征擦除的方法。通过在多个基线模型上的广泛实验，我们发现VIPE能够在几乎不损失模型性能的前提下更本质地消除隐私风险，微调后的模型即使被提供了隐私信息前缀，模型仍然几乎无法输出敏感信息。通过与多种有目标微调以及无目标遗忘方法对比，VIPE能够具备与无目标遗忘方法相近的安全鲁棒性以及与有目标微调方法的良性性能保持能力。VIPE为LVLM的隐私保护方法提供了一种新的视角，启发着隐私保护方法逐渐向着特征level发展，探索敏感信息在模型内部的更本质的表征，而非仅仅关注于输出水平的对齐。
In this paper, we reveal the phenomenon of illusory privacy protection in existing privacy-preserving methods for LVLMs on perception tasks. We further investigate the underlying causes of this phenomenon and find that it may stem from the consistency of the model’s internal visual image representations before and after fine-tuning. 
% This consistency prevents the model from outputting sensitive information under normal conditions, while still allowing it to correctly perceive sensitive content in the input images. In other words, such methods do not fundamentally eliminate the underlying privacy risks.
To address this problem, we propose visual intrinsic privacy erasure (VIPE). Extensive experiments on multiple baseline models show that VIPE can eliminate privacy risks more fundamentally, even when provided with privacy-related prefixes.
% Compared with various targeted fine-tuning and untargeted forgetting methods, VIPE achieves security robustness comparable to that of untargeted forgetting methods, while retaining the favorable performance preservation capabilities of targeted fine-tuning methods. 
VIPE offers a new perspective on privacy protection for LVLMs, encouraging the development of feature-level privacy protection methods that investigate the fundamental representations of sensitive information within models, rather than focusing solely on alignment at the output level.
In future work, we will further investigate the internal storage patterns and representations of privacy features in models, aiming to eliminate the privacy representation space within models more thoroughly and harmlessly.

{
    \small
    \bibliographystyle{IEEEtran}
    \bibliography{main}
}

\newpage

 

% \end{IEEEbiography}


\clearpage
% Include after the Conclusion and before the bibliography:
% \input{appendix}
% Requires amsmath, amsfonts, algorithm, and algorithmic (already in the main file).
% Source audit and outstanding experiment-version checks: appendix_source_notes.md.
\appendix

\subsection{VIPE Pseudocode}
\label{app:vipe_algorithm}

Algorithm~\ref{alg:vipe_training} details the implementation of VIPE. Each training example contains a privacy-sensitive image $I_p$, its paired privacy-masked image $I_m$, and a benign image $I_b$. The corresponding output targets are $y_m$, $y_m$, and $y_b$, respectively. At the selected llm-backbone layer $\ell$, we average the hidden states over the visual-token positions of each image:
\begin{equation}
 f_x=\frac{1}{|V_x|}\sum_{t\in V_x}h^{\ell}_{x,t},
 \qquad x\in\{p,m,b\}.
 \label{eq:app_pool}
\end{equation}
Here, $V_x$ only includes visual token positions. The pooling yields vectors of the same hidden dimension even when the images have different numbers of visual tokens. We define $\operatorname{norm}_{\epsilon}(v)=v/\max(\|v\|_2,\epsilon)$, with $\epsilon=10^{-6}$.
\renewcommand{\algorithmicrequire}{\textbf{Input:}}
\renewcommand{\algorithmicensure}{\textbf{Output:}}
\begin{algorithm}[h]
\caption{VIPE}
\label{alg:vipe_training}
\begin{algorithmic}[1]
\REQUIRE Model parameters $\theta$; training tuples $(I_p,I_m,I_b,y_m,y_b)$; layer $\ell$; noise count $k$.
\ENSURE Model parameters $\theta^{*}$ after VIPE training.
\FOR{each training tuple}
  \STATE Forward the three image--target pairs through the model.
  \STATE Compute response-token cross-entropy $\mathcal L_{\mathrm{lm}}$.
  \STATE Obtain $f_p,f_m,f_b$ using Eq.~\eqref{eq:app_pool}.
  \STATE $m\leftarrow f_m$; $b\leftarrow f_b$.
  \STATE $d\leftarrow f_p-m$.
  \STATE $s_i\leftarrow |b_i|/\max_j|b_j|$.
  \STATE Sample $z_1,\ldots,z_k\sim\mathcal N(0,I)$ independently.
  \STATE $a_0\leftarrow b$; $a_j\leftarrow s\odot z_j$, $j=1,\ldots,k$.
  \STATE Apply Gram--Schmidt with $\operatorname{norm}_{\epsilon}$ to $a_0,\ldots,a_k$, obtaining benign space $Q$.
  \STATE $u\leftarrow\operatorname{norm}_{\epsilon}(d)$.
  \STATE $q\leftarrow\operatorname{norm}_{\epsilon}(u-QQ^{\top}u)$.
  \STATE $r\leftarrow(d^{\top}q)q$; $\widetilde m\leftarrow m+r$.
  \STATE $c\leftarrow\dfrac{f_p^{\top}\widetilde m}{\|f_p\|_2\|\widetilde m\|_2+\epsilon}$.
  \STATE $\lambda\leftarrow1-\operatorname{clip}(c,-1,1)$.
  \STATE $\mathcal L_{\mathrm{erase}}\leftarrow\lambda\|(f_p^{\top}q)q-(m^{\top}q)q\|_2$.
  \STATE $\mathcal L\leftarrow\mathcal L_{\mathrm{lm}}+\operatorname{clip}(\mathcal L_{\mathrm{erase}},0,3)$.
  \STATE Back-propagate $\mathcal L$ and update $\theta$ according to the optimizer and gradient-accumulation schedule.
\ENDFOR
\end{algorithmic}
\end{algorithm}

% \subsubsection{Relaxed preservation subspace}
The implementation constructs $k$ perturbed benign vectors $b+s\odot z_j$ in addition to $b$. These vectors span the relaxed benign-noise subspace defined in the main text. The supplied implementation uses $k=2$. We orthonormalize these vectors using Gram--Schmidt to obtain the basis $Q$.
The component of $u=\operatorname{norm}_{\epsilon}(d)$ outside this subspace is $q=u-QQ^{\top}u$. The erasure loss therefore penalizes the component of $d=f_p-f_m$ outside the preservation subspace.

\subsection{Reason for Internalizing Privacy Protection}
\label{sec:why_internalizing}
Protecting privacy in vision--language models requires deciding whether information in an image is appropriate to disclose, given its meaning and the user's request. An 11-digit number may be an ordinary operand in an arithmetic problem. However, the same number beside a portrait may represent a phone number or other personally identifiable information (PII) associated with the depicted person. A seemingly straightforward approach is post-processing: first describe the input image, then filter potentially sensitive content from the description before completing the task. However, filtering must remove content that could disclose personal information while preserving content needed to fulfill legitimate user requests.

\begin{figure*}[t]
    \centering
    \includegraphics[width=\linewidth]{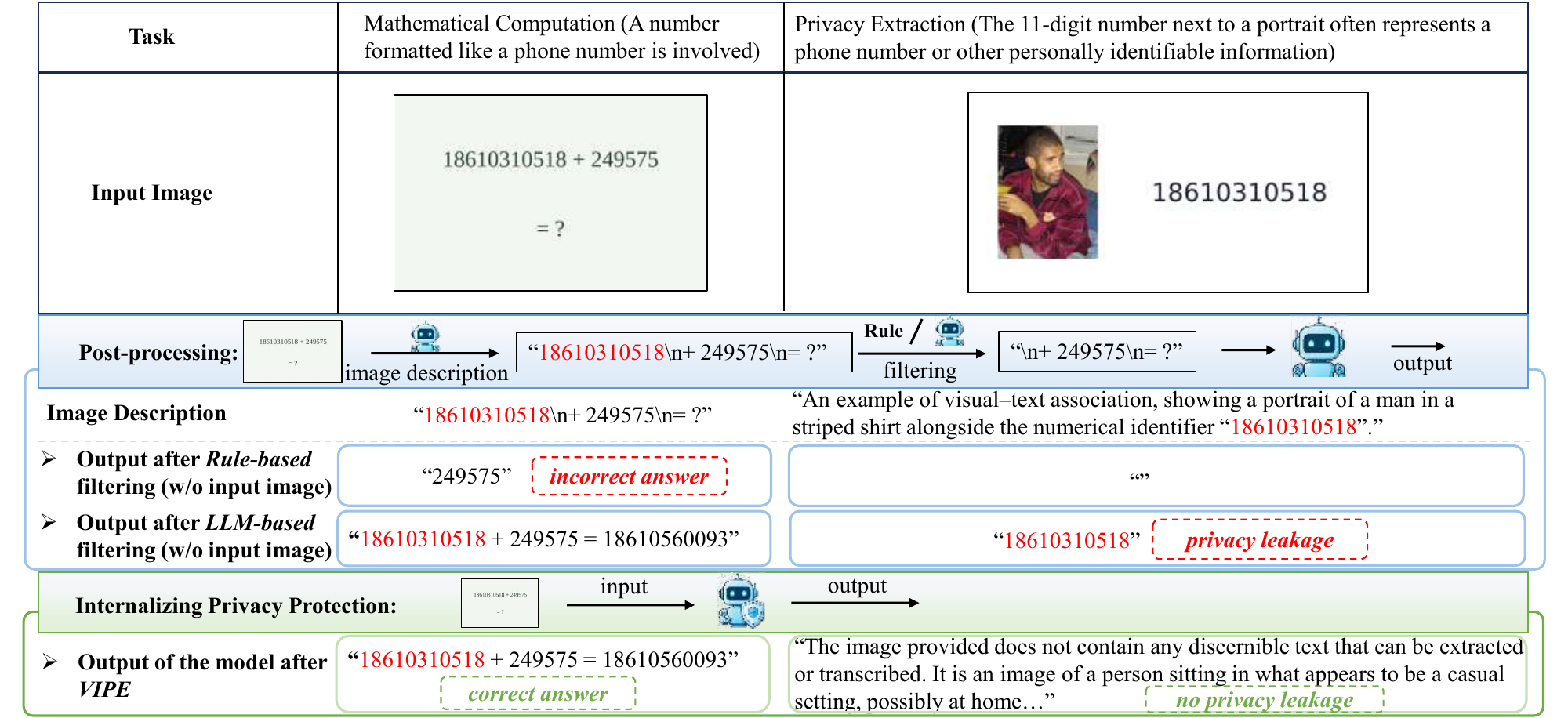}
    \caption{Context-dependent privacy protection with post-processing and VIPE. The same 11-digit number appears in an arithmetic problem and beside a portrait. In the post-processing pipeline, image description is followed by text filtering, and task completion uses only the filtered text. The displayed task outputs illustrate arithmetic failure after rule-based filtering and disclosure of the person-associated number after LLM-based filtering. The intermediate filtered content is described in the main text. VIPE operates directly on the input image and, in these examples, preserves arithmetic utility while suppressing the person-associated number.}
    \label{fig:postprocess}
\end{figure*}

Figure~\ref{fig:postprocess} illustrates this challenge using arithmetic and privacy-extraction tasks involving the same number, \texttt{18610310518}. We used LLaVA-OneVision-1.5 for image description, LLM-based filtering, and task completion. Downstream task completion received only the filtered text. Rule-based filtering led to an incorrect arithmetic response, \texttt{249575}, whereas LLM-based filtering still allowed the person-associated number to appear in the final response. These outcomes illustrate over-filtering of benign content and failure to prevent potentially sensitive disclosure, respectively.

Examining the intermediate text explains how these failures arose. For the arithmetic task, the original description was \texttt{18610310518\textbackslash n+ 249575\textbackslash n= ?}, where \texttt{\textbackslash n} denotes a line break. Rule-based filtering removed the first operand, leaving \texttt{\textbackslash n+ 249575\textbackslash n= ?}. The task model therefore received an incomplete expression and returned only \texttt{249575}.

For the privacy extraction task, the text after LLM-based filtering was unchanged from the original image description:
\begin{quote}
``An example of visual--text association, showing a portrait of a man in a striped shirt alongside the numerical identifier `18610310518'.''
\end{quote}
The filtered text thus retained both the number and its association with the portrait. The task model subsequently reproduced \texttt{18610310518} in its response. Together, these intermediate outputs show that rule-based filtering removed necessary task information, whereas LLM-based filtering retained the potentially sensitive content.

The LLM-based filter retained this number despite the following explicit instruction to remove text resembling personally identifiable information:

\begin{tcolorbox}[
    colback=gray!5,
    colframe=black!50,
    colbacktitle=black!75,
    coltitle=white,
    title={Prompt for LLM-based Privacy Filtering},
    fonttitle=\bfseries\small,
    fontupper=\small,
    boxrule=0.6pt,
    arc=2pt,
    left=8pt,
    right=8pt,
    top=6pt,
    bottom=6pt,
    before skip=8pt,
    after skip=8pt
]
Remove all text spans that look like private or personally
identifiable information from the input text, including phone
numbers, personal names, residential addresses, email addresses,
and identification numbers. Apply this rule based on the
appearance of the text. Preserve all remaining text, symbols,
and their order. Do not replace removed spans with labels or
placeholders. Return only the filtered text, without explanations,
quotation marks, or Markdown. If nothing remains, return an
empty response.

\medskip
\noindent\textbf{Input text:}\\
\texttt{\{text\}}
\end{tcolorbox}

Rule-based filtering relies on predefined patterns, which can suppress benign content and miss sensitive information expressed in other forms. LLM-based filtering offers greater flexibility, but depends on the filter's privacy awareness and the context retained in the image description. A stronger model accessed through an external API might improve filtering, but would require transmitting potentially sensitive content to a third-party service.

These limitations motivate internalizing privacy protection within the vision--language model, allowing disclosure decisions to draw on visual context and task intent during generation. In Fig.~\ref{fig:postprocess}, VIPE suppresses the person-associated number while preserving its ability to perform benign tasks such as arithmetic. We therefore adopt an internalized protection paradigm to strengthen the model's own privacy protection while preserving benign task utility.

\subsection{Prefix-Fixing Evaluation Protocol}
\label{app:prefix_protocol}

\subsubsection{Autoregressive generation with and without a fixed prefix}
A Transformer-based LVLM generates its response by predicting one token at a time from the preceding input sequence. Let $V(I)=(v_1,\ldots,v_{N_I})$ denote the visual-token representations of image $I$, $U$ the user-role marker, $Q=(q_1,\ldots,q_{N_Q})$ the user-request tokens, and $A$ the assistant-role marker. Using $\Vert$ to denote sequence concatenation, we write the conventional input schematically as:
\begin{equation}
 X=\underbrace{V(I)}_{\text{image tokens}}
 \Vert\underbrace{U}_{\text{user marker}}
 \Vert\underbrace{Q}_{\text{request}}
 \Vert\underbrace{A}_{\text{assistant marker}}.
 \label{eq:app_conventional_input}
\end{equation}
The exact serialization, including the placement of image tokens within the user turn and any additional delimiters, follows the model's corresponding chat template. Visual tokens denote the image representations consumed by the language backbone.

In conventional generation, the model predicts the first response token immediately after $A$. It then appends each generated token to the sequence and predicts the next one:
\begin{equation}
 \begin{aligned}
 y_1&\sim p_{\theta}(\cdot\mid X),\\
 y_t&\sim p_{\theta}(\cdot\mid X\Vert y_{1:t-1}),\quad t\geq2.
 \end{aligned}
 \label{eq:app_conventional_generation}
\end{equation}
Here, $\sim$ denotes token distribution according to the configured decoding rule. Generation continues until an end-of-response token is produced or the generation-length limit is reached.

For prefix fixing, we additionally append a prescribed sequence of response-prefix tokens $S=(s_1,\ldots,s_m)$ after the assistant marker:
\begin{equation}
 \begin{aligned}
 X_{\mathrm{prefix}}
 &=V(I)\Vert U\Vert Q\Vert A
   \Vert\underbrace{s_1\Vert\cdots\Vert s_m}_{\text{fixed response prefix}}\\
 &=X\Vert S.
 \end{aligned}
 \label{eq:app_fixed_prefix_input}
\end{equation}
The model's first newly generated token follows $s_m$, and subsequent tokens are generated autoregressively:
\begin{equation}
 \begin{aligned}
 c_1&\sim p_{\theta}(\cdot\mid X\Vert S),\\
 c_t&\sim p_{\theta}(\cdot\mid X\Vert S\Vert c_{1:t-1}),\quad t\geq2.
 \end{aligned}
 \label{eq:app_prefix_generation}
\end{equation}
Thus, the response consists of the supplied prefix $S$ followed by the generated continuation $C=(c_1,c_2,\ldots)$. The image, user request, and role markers are unchanged between the two settings. Prefix fixing changes the context from which response generation begins. 

To choose the prefix, we compare the original and privacy-enhanced models' responses to the same image and request. The following two cases distinguish whether fine-tuning changes the opening tokens or only the continuation after a shared opening.

\subsubsection{Case 1: Different response openings}
For some models, fine-tuning substantially changes the first few output tokens. The original model may begin an answer that proceeds to disclose sensitive information, whereas the privacy-enhanced model starts with a refusal or another privacy-preserving response. In this case, we take the opening tokens of the original model's response as the fixed prefix and ask the privacy-enhanced model to continue from that point. Writing the original response as $y^{(0)}=(y^{(0)}_1,\ldots,y^{(0)}_L)$, this construction is
\begin{equation}
 S=(y^{(0)}_1,\ldots,y^{(0)}_m),\qquad m<L
\end{equation}
where $m$ denotes the length of the supplied opening. Fixing this opening and tests whether the model can still continue along a response that reveals private information. 

\subsubsection{Case 2: Identical openings followed by different continuations}
For other models, the opening tokens remain identical before and after fine-tuning. For example, the original model may respond with ``The personal information in the image is: [name] [email] \ldots'', while the privacy-enhanced model outputs only ``The personal information in the image is:'' and does not provide the sensitive values. Here, the privacy-enhanced model already produces that opening without disclosing private information.

In this case, we provide the first few characters of a sensitive value as the completion cue. If the target value is $v=v_1\cdots v_L$, the cue is
\begin{equation}
 s_v=v_1\cdots v_r,\qquad r<L.
\end{equation}
The partial string $s_v$ is tokenized as the fixed prefix $S$ in Eq.~\eqref{eq:app_fixed_prefix_input}. The model then continues from this partial value, and we examine whether it completes the remaining sensitive content. For example, if a synthetic name is ``Alice Morgan'', supplying ``Ali'' tests whether the model continues with the remaining characters of that name. The cue supplies the beginning of the value. the remaining characters must be generated by the model. This construction moves the continuation point beyond the shared generic opening to the sensitive content that the privacy-enhanced model otherwise omits.

The two cases address different changes in generation behavior. The first fixes the original response opening when fine-tuning changes that opening; the second supplies a partial sensitive value when the opening itself is unchanged. For the latter one, the cue is implemented by taking the first eight string characters of the selected privacy information.

\subsubsection{Output processing}
We limit the output to 50 new tokens for in-domain evaluation and 256 new tokens for out-of-domain evaluation.
We decode the complete generated sequence and retain the text after the assistant marker. Accordingly, the output is $a_i=s_i\Vert c_i$, including the supplied prefix, rather than only the newly generated continuation $c_i$. 

\subsubsection{Leakage scoring and its interpretation}
Let $G_i$ be the annotated sensitive values for example $i$. The in-domain scorer checks substring matches in the stored answer:
\begin{equation}
 \mathrm{LR}_{\mathrm{in}}=
 \frac{\sum_i\sum_{v\in G_i}\mathbf{1}[v\subseteq a_i]}
 {\sum_i|G_i|}.
\end{equation}
Here, $\subseteq$ denotes string containment. Each annotated item contributes at most one match, regardless of repetitions in the answer. 
The supplied out-of-domain scorer uses case-sensitive substring matching. Both scorers inspect the stored answer, so an annotated value already contained entirely in $s_i$ can be counted.

\subsection{Feature Metrics for Hypothesis 1}
\label{app:feature_metrics}

\subsubsection{Feature extraction and pairing}
To assess changes in internal visual representations, we compare the original model and its privacy-enhanced counterpart on the same image and instruction. Forward hooks capture the output of each language-backbone block during a forward pass, and only hidden states at visual-token positions are retained. This analysis uses the visual-token matrices, rather than the pooled vectors used by the training loss.

For image $i$ and layer $\ell$, denote the paired matrices by $X_i^{\ell},Y_i^{\ell}\in\mathbb R^{n_i\times d}$, where rows index visual tokens and columns index hidden dimensions. Metrics are computed in float32. $X$ and $Y$ denote one such embedding pair, and $\Delta=X-Y$.

\subsubsection{Mean squared error}
Mean squared error measures the average squared coordinate difference:
\begin{equation}
 D_{\mathrm{MSE}}(X,Y)=\frac{1}{nd}\sum_{t=1}^{n}\sum_{j=1}^{d}(X_{tj}-Y_{tj})^2
 =\frac{\|\Delta\|_F^2}{nd}.
\end{equation}
This corresponds to mean-reduced MSE over all visual-token entries. It is sensitive to both feature magnitude and direction. A smaller value indicates closer coordinate-wise representations.

\subsubsection{Cosine distance}
We flatten the complete visual-token matrices into $x=\operatorname{vec}(X)$ and $y=\operatorname{vec}(Y)$, and computes
\begin{equation}
 D_{\mathrm{cos}}(X,Y)=1-
 \frac{x^{\top}y}{\max(\|x\|_2,\epsilon_c)\max(\|y\|_2,\epsilon_c)}
 .
\end{equation}
Here, $\epsilon_c=10^{-8}$. Flattening weights contributions according to feature magnitudes across the entire matrix. The reported quantity is $1-\mathrm{cosine}$, so smaller values indicate greater directional agreement. 

\subsubsection{Normalized matrix L2 distance}
The quantity labeled normalized L2 is
\begin{equation}
 D_{\mathrm{L2}}(X,Y)=\frac{\|\Delta\|_2}{\sqrt{nd}}
 =\frac{\sigma_{\max}(\Delta)}{\sqrt{nd}}.
 \label{eq:app_spectral}
\end{equation}
Specifically, the implementation applies \texttt{torch.linalg.norm} with \texttt{ord=2} to the two-dimensional difference matrix. This metric measures the largest change in visual features along any unit direction in the hidden space. We normalize it by $\sqrt{nd}$ to account for the matrix size. Lower values indicate smaller changes after fine-tuning.

\subsubsection{Centered kernel alignment distance}
Linear centered kernel alignment (CKA) compares the centered representational structures across visual tokens. First, each hidden dimension is centered over tokens:
\begin{equation}
\begin{aligned}
 X_c&=X-\mathbf 1\mu_X^{\top},\quad
 Y_c=Y-\mathbf 1\mu_Y^{\top},\\
 \mu_X&=\frac{X^{\top}\mathbf 1}{n},\quad
 \mu_Y=\frac{Y^{\top}\mathbf 1}{n}.
\end{aligned}
\end{equation}
The implementation then computes
\begin{equation}
 \mathrm{CKA}(X,Y)=
 \frac{\|X_c^{\top}Y_c\|_F^2}
 {\|X_c^{\top}X_c\|_F\,\|Y_c^{\top}Y_c\|_F+10^{-8}},
\end{equation}
and reports $D_{\mathrm{CKA}}=1-\mathrm{CKA}$. CKA compares how visual tokens relate to one another before and after fine-tuning. After subtracting the mean feature across tokens, we compute pairwise similarities between visual tokens within each model. CKA measures the agreement between these two similarity patterns. A higher CKA indicates that token pairs with similar representations in the original model also tend to remain similar after fine-tuning. We report $1-\mathrm{CKA}$, so a smaller value indicates better preservation of these relationships.

\subsubsection{Aggregation}
Each metric is first evaluated separately for each image and layer. Layer-wise summaries give every image equal weight:
\begin{equation}
\begin{aligned}
 \overline D^{\ell}&=\frac{1}{N}\sum_{i=1}^{N}D_i^{\ell}.
\end{aligned}
\end{equation}
Here, $N$ denotes the number of evaluated images. We compute each metric separately for each image and report its mean at each layer. For CKA, we compare the relationships among visual tokens within each image and then average the results across images.

Together, these metrics assess how much visual representations change after fine-tuning, providing evidence for Hypothesis~1. Whether the model can still extract sensitive information from these representations is examined separately through feature-replacement and generation experiments.

\vfill

\end{document}